\documentclass[letterpaper]{article} 
\usepackage{aaai2026}  
\usepackage{times}  
\usepackage{helvet}  
\usepackage{courier}  
\usepackage[hyphens]{url}  
\usepackage{graphicx} 
\usepackage{natbib}  
\usepackage{caption} 
\usepackage{algorithm}
\usepackage{algorithmic}
\usepackage{amsmath}
\usepackage{amsthm}
\usepackage{amssymb}
\usepackage{multirow}
\usepackage{booktabs}
\usepackage{diagbox}
\usepackage{pifont}

\newtheorem{theorem}{Theorem}
\newtheorem{lemma}{Lemma}
\newtheorem{definition}{Definition}

\usepackage{newfloat}
\usepackage{listings}
\DeclareCaptionStyle{ruled}{labelfont=normalfont,labelsep=colon,strut=off} 
\floatstyle{ruled}
\newfloat{listing}{tb}{lst}{}
\floatname{listing}{Listing}
\title{Tighter Truncated Rectangular Prism Approximation \\for RNN Robustness Verification}
\author{
    Xingqi Lin\textsuperscript{\rm 1}, Liangyu Chen\textsuperscript{\rm 1}\thanks{Liangyu Chen is the corresponding author.}, Min Wu\textsuperscript{\rm 1}, Min Zhang\textsuperscript{\rm 1}, Zhenbing Zeng\textsuperscript{\rm 2}
}
\affiliations{
    \textsuperscript{\rm 1}Shanghai Key Laboratory of Trustworthy Computing\\
    \textsuperscript{\rm 2}Department of Mathematics, Shanghai University\\
    lychen@sei.ecnu.edu.cn

}

\usepackage{bibentry}

\begin{document}

\maketitle

\begin{abstract}
Robustness verification is a promising technique for rigorously proving Recurrent Neural Networks (RNNs) robustly. A key challenge is to over-approximate the nonlinear activation functions with linear constraints, which can transform the verification problem into an efficiently solvable linear programming problem. Existing methods over-approximate the nonlinear parts with linear bounding planes individually, which may cause significant over-estimation and lead to lower verification accuracy. In this paper, in order to tightly enclose the three-dimensional nonlinear surface generated by the Hadamard product, we propose a novel truncated rectangular prism formed by two linear relaxation planes and a refinement-driven method to minimize both its volume and surface area for tighter over-approximation. Based on this approximation, we implement a prototype \emph{DeepPrism} for RNN robustness verification. The experimental results demonstrate that \emph{DeepPrism} has significant improvement compared with the state-of-the-art approaches in various tasks of image classification, speech recognition and sentiment analysis.
\end{abstract}

\section{Introduction}
The widespread application of artificial intelligence has raised growing concerns about its security. This is particularly critical in scenarios with low fault tolerance, such as obstacle detection for autonomous driving and diagnostic classification in medical imaging, where neural network errors can lead to catastrophic consequences. Some studies \cite{su2019one} even show that one single pixel attack can fool neural networks, exposing the vulnerability to adversarial attacks. Nevertheless, due to the black-box essentiality and vast scale of neural networks, it is not practical to evaluate their security by exhaustively enumerating all possible inputs. Therefore, constructing an effective verification framework to analyze the robustness of these networks is important and necessary.

The problem of neural network robustness verification can be described as follows: Given an input $x$ and a perturbation $\epsilon$, does the result remain consistent with the original $x$ or within an acceptable margin of error? For classification tasks, verification typically involves checking whether the predicted probability of the target class stays higher than that of other classes after perturbation. Current researches have mainly focused on the verification of Feedforward Neural Networks (FNNs), with relatively less attention on Recurrent Neural Networks (RNNs). This restricts the full potentiality of RNNs in vital applications.

RNN is a type of artificial neural network designed to process sequential data. Its architecture maintains a memory of past inputs, making it well-suited for tasks like natural language processing, speech recognition, and time series forecasting. However, their susceptibility to adversarial examples becomes an increasing worry \cite{papernot2016crafting}. The key challenge in RNN verification is the nonlinearity of the activation and gated functions. Typically, this issue is tackled by over-approximating the initial network to construct a linear problem with relaxed abstract domains. Taking the classic RNN Long Short-Term Memory (LSTM) as an example, nonlinear operations like $\sigma(x) \odot \tanh(y)$, which involve the multiplication of two variables, significantly increase the verification burden. The previous work \cite{ryou2021scalable} applies DeepPoly \cite{singh2019abstract} to RNN verification, using Linear Programming (LP) to obtain the upper and lower planes of gated functions. The objective function of LP is the sum of the vertical distances between the surface and the planes at sampled points. By minimizing it, one can obtain planes that are closer to the surface. While this method is intuitive, it neglects the relationship between the bounding planes, thus losing verification accuracy.

In this paper, we focus on the relaxation problem of $\sigma (x) \odot \tanh(y)$, analyze the truncated rectangular prism formed by two linear relaxation planes and propose a tighter relaxation method based on the hybrid objective function of its volume and surface area. Through strict mathematical deduction, we demonstrate that the height of the centroid between the upper and lower planes is proportional to the volume, while the surface area is positively correlated with the difference between the maximum value of the upper plane and the minimum value of the lower plane. Therefore, we obtain a more straightforward and theoretical approximation method by using the weighted sum of the height and the difference as the optimization objective. Based on this, we design and implement a RNN verifier called \emph{DeepPrism}, which yields a notable improvement on robustness verification in various tasks of image classification, speech recognition and sentiment analysis, outperforming previous work.

Our main contributions are as follows:
\begin{itemize}
     \item We introduce the truncated rectangular prism formed by two relaxation planes and minimize its volume and surface area to achieve a tighter over-approximation, thereby constructing effective abstract domains for RNN robustness verification.
     \item We propose an over-approximation approach that combines linear programming with a hybrid objective function and abstraction refinement based on different division strategies.
    \item We implement our approach as a RNN verifier \emph{DeepPrism}, and evaluate it through experiments. The experimental results on four datasets for three tasks show that \emph{DeepPrism} outperforms other SOTA baselines with higher  accuracy. The code and data are available in https://github.com/Olinvia/DeepPrism .
\end{itemize}

We describe our related work, preliminary, methodology, experiments and conclusion in the following sections. Detailed proofs and more results are provided in the appendix.

\section{Related Work}
\label{related}
Methods for neural network verification can be categorized into exact methods and approximate methods. Exact methods mainly include Satisfiability Modulo Theory (SMT) \cite{katz2017reluplex,katz2019marabou,dpll,isac2025proof} and Mixed Integer Linear Programming (MILP) \cite{bunel2018unified,dutta2018output,Xue_2022_ACCV}. They can precisely compute the reachable sets of the neural network output but suffer from high complexity, heavy computational cost, and limited scalability. Since around 2018, approximate methods have gradually gained prominence due to their efficiency. Approximate methods include abstract interpretation \cite{gehr2018ai2,singh2018fast,lemesle2024neural,marzari2025advancing}, symbolic propagation \cite{wang2018formal,wang2018efficient,pmlr-v270-hu25a}, and convex optimization \cite{muller2022prima, wu2022efficient}, etc. They offer significant advantages in computational efficiency and scalability, making them applicable to larger-scale neural networks and more complex application scenarios.

For RNN verification\cite{diffRNN,SafeDeep}, there are three mainstream approximate methods: abstract interpretation, RNN2FNN-based verification, and automata-based methods. Abstract interpretation maps the internal structure of RNN to a set of specific geometric shapes and then verifies whether the abstract domain of the output layer satisfies robustness properties. It is efficient and scalable, but the approximation may lead to a loss of accuracy. RNN2FNN-based verification involves converting the inputs at all time steps into static inputs and applying the verification methods used for FNNs. The methods balance reliability and completeness but come with higher costs. The automata-based method extracts an automaton or finite-state machine from the RNN, which requires a higher level of theoretical knowledge. 

Since this paper focuses on the abstract interpretation methods, we summarize them as follows. 

\noindent (1) Ko et al. \shortcite{pmlr-v97-ko19a}, inspired by Fastlin \cite{weng2018towards}, which adds linear constraints to neural network operations, first applied abstract interpretation to RNN verification and proposed \emph{POPQORN}. 
The nonlinear parts are bounded by linear functions, which can be propagated back to the first layer from the output layer recursively. 
However, this method can handle only a limited number of neurons and may result in overly loose robustness bounds.

\noindent (2) Based on \emph{POPQORN}, Du et al .\shortcite{du2021cert} applied the ideas from DeepZ \cite{singh2018fast} to RNN verification and proposed a tighter verification framework, \emph{Cert-RNN}. Their relaxation strategy for S-shaped activation functions such as sigmoid and tanh is more precise, leveraging the properties of tangents. Additionally, they refined the linear bounds for Hadamard products by conducting a case-by-case analysis, achieving better experimental results than \emph{POPQORN}.

\noindent (3) Ryou et al. \shortcite{ryou2021scalable} proposed a new RNN verifier, \emph{Prover}. They drew on the ideas of DeepPoly and introduced numerical and symbolic bounds for each neuron. For each layer, the bounds are propagated back to the input layer. Moreover, their relaxation approach employs linear programming to find the upper and lower bounding planes individually. 

\noindent (4) Zhang et al. \shortcite{zhang2023rnnguard} proposed \emph{RNN-Guard}, a certified defense against multi-frame attacks for RNNs. They designed an abstract domain called InterZono, which achieves twice the verification precision compared to the Zonotope \cite{zonotope2009}.

\section{Preliminaries}
\label{bg}
\subsection{LSTM}
\begin{figure}[ht]
    \centering
    \includegraphics[width=0.4\textwidth]{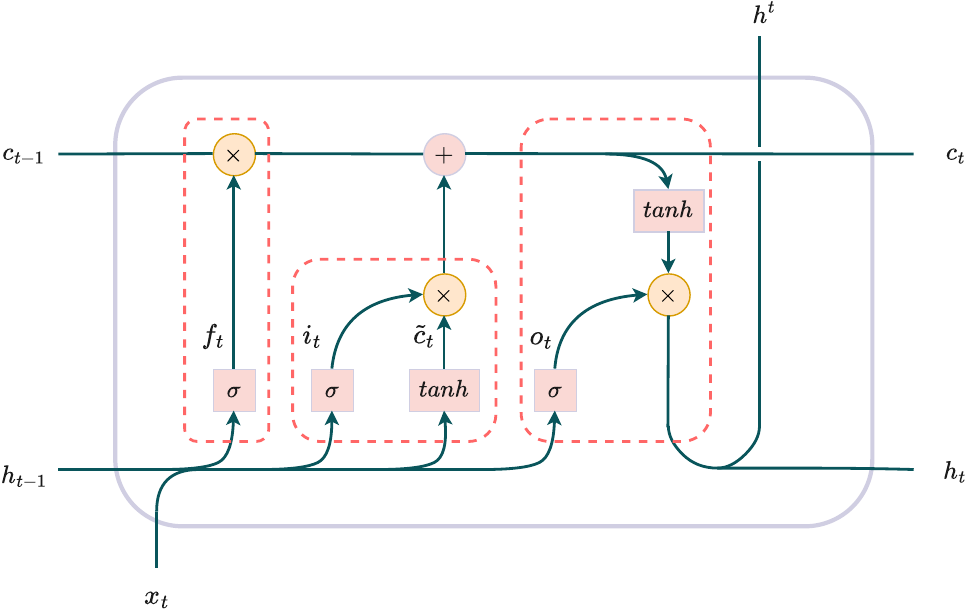}
    \caption{An LSTM cell consists of a forget gate, an input gate, and an output gate. The regions highlighted by the red dashed line indicate the key challenge of over-approximation of $\sigma(x) \odot \tanh(y)$ and $\sigma(x) \odot y$.}
    \label{lstm}
\end{figure}

LSTM is a type of RNN, whose architecture is shown in Fig.~\ref{lstm}. It has three gate functions explained as follows. 

The forget gate determines what information should be discarded from the cell state and is represented as follows: 

\begin{equation}
    f_t = \sigma (W_f \cdot [h_{t-1},x_t]+b_f),
\end{equation}
where $\sigma (x) = \frac{1}{1+e^{-x}}$ is the activation function, $W_f$ and $b_f$ are the weight and bias, and $[h_{t-1},x_t]$ represents the concatenation of the previous hidden state and the current input. 

The input gate controls what new information should be added to the cell state and decides the extent to which the cell's memory is updated. It can be expressed as:
\begin{align}
    i_t &= \sigma (W_i \cdot [h_{t-1},x_t]+b_i), \tag{2} \label{eq:it} \\
    \tilde{c}_t &= \tanh(W_C \cdot [h_{t-1},x_t]+b_C), \tag{3} \label{eq:ct}
\end{align}
where $\tanh(x) = \frac{e^x-e^{-x}}{e^x+e^{-x}}$ is the activation function, $i_t$ is the output, $\tilde{c}_t$ is the candidate memory cell state, and $W_i, W_C$ and $b_i, b_c$ are the weights and biases, respectively.

The output gate decides what the next hidden state should be based on the current cell state:
\begin{align}
    o_t &= \sigma (W_o \cdot [h_{t-1},x_t]+b_o), \tag{4} \label{eq:ot} \\
    c_t &= f_t \odot c_{t-1} +i_t \odot \tilde{c}_t, \tag{5} \label{eq:cct} \\
    h_t &= o_t \odot \tanh(c_t), \tag{6} \label{eq:ht}
\end{align}
where $\odot$ is Hadamard product, $o_t$ is the output, $c_t$ is the cell state, and $h_t$ is the hidden state. The nonlinearity of Eq.~5 and Eq.~6, such as $\odot$ operation, is the challenge of RNN verification.

\subsection{DeepPoly}
\label{DeepPoly}
DeepPoly \cite{singh2019abstract} is an effective framework for FNN verification. Let the set of neurons at layer $\ell$ in the network be $X^{(\ell)} = \{x_1^{(\ell)}, x_2^{(\ell)}, \ldots, x_n^{(\ell)}\}$. For each neuron $x_i^{(\ell)}$, there are numerical constraints $l_i \leq x_i^{(\ell)} \leq u_i$ and symbolic constraints $a_i \cdot x_j^{(\ell-1)} + b_i \leq x_i^{(\ell)} \leq a_i^{\prime} \cdot x_j^{(\ell-1)} + b_i^{\prime}$, where $l_i, u_i, a_i, b_i, a_i^{\prime}, b_i^{\prime} \in \mathbb{R}$. Nonlinear neurons at layer $\ell$ can be linearly over-approximated by neurons at layer $\ell-1$, and this process can be recursively traced back to the first layer. Therefore, given the input and perturbation, the output range can be calculated.

\section{Tighter Over-approximation}
\label{tighter}
DeepPoly demonstrates that as long as symbolic linear bounds for nonlinear functions can be found, the output range can be determined by using backsubstitution. Thus, the core problem of verification is finding appropriate linear approximation methods. Here, we take $\sigma(x) \odot \tanh(y)$ as an example to discuss how we obtain tighter abstraction domains in abstract interpretation.

\subsection{Distance-based Method \cite{ryou2021scalable}}
\label{distance}

We first briefly introduce the distance-based method called \emph{Prover} \cite{ryou2021scalable}. LSTM involves two multiplications that require approximation, and we use $f(x, y)=\sigma (x) \odot \tanh(y)$ as an example. We need to find the upper and lower bounding planes of $f$ such that:
$A_l \cdot x+B_l \cdot y+C_l \leq f(x,y) \leq A_u\cdot x+B_u \cdot y+C_u.$
It can be transformed as an optimization problem, namely, the variables are the coefficients of the planes $A_l$, $B_l$, $C_l$, $A_u$, $B_u$, $C_u$, the constraints should ensure that the upper bounding plane always lies above the surface and the lower bounding plane always lies below the surface, and the objective function defines how we evaluate the quality of the abstract domain. Ryou et al. proposed minimizing the vertical distance between the surface and the planes, which are expressed as:

\begin{equation}
    \begin{split}
        \min \limits_{A_l,B_l,C_l}{\int_{(x,y) \in B }} (f(x,y)-(A_l \cdot x + B_l \cdot y + C_l)) \\
        \text{s.t.} \; A_l \cdot x + B_l \cdot y + C_l \leq f(x,y), \forall (x,y) \in B,
    \end{split}
    \tag{7}
\end{equation}
and
\begin{equation}
    \begin{split}
        \min \limits_{A_u,B_u,C_u}{\int_{(x,y) \in B }} ((A_u \cdot x + B_u \cdot y + C_u)-f(x,y)) \\
        \text{s.t.} \; A_u \cdot x + B_u \cdot y + C_u \geq f(x,y), \forall (x,y) \in B.
    \end{split}
    \tag{8}
\end{equation}
where $B$ is $[l_x,u_x] \times [l_y,u_y]$.

\begin{figure}[ht]
    \centering
    \includegraphics[width=0.4\textwidth]{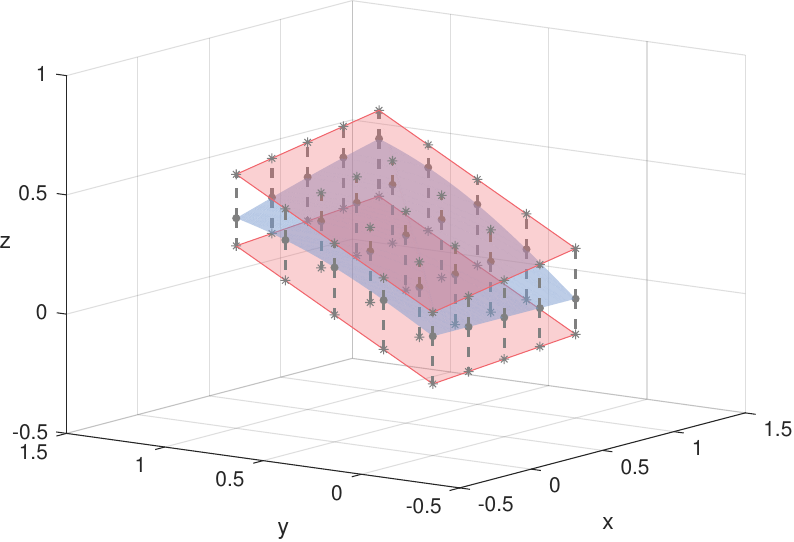}
    \caption{The upper and lower planes computed by linear programming based on the distance sum of sampling points.}
    \label{object-a}
\end{figure}

Eq.~7 and Eq.~8 are implemented by sampling, essentially representing the surface features by several points, as shown in Fig.~\ref{object-a}. The constraints are satisfied at the $n$ sampled points, and the objective function is the sum of the distances at these $n$ points. The more points sampled, the more precise the approximation, but the time cost increases accordingly. However, sampling cannot fully guarantee soundness, such as ensuring that the lower bounding plane is always beneath the surface. Therefore, after the linear programming is completed and $A_l$, $B_l$, and $C_l$ are obtained, it is necessary to check whether the curve surface and the lower plane intersect. If they do, the planes should be adjusted with offsets. The offset algorithm can be referred in \cite{ryou2021scalable}.

\subsection{Volume-based Method}
\begin{figure}[ht]
    \centering
    \includegraphics[width=0.4\textwidth]{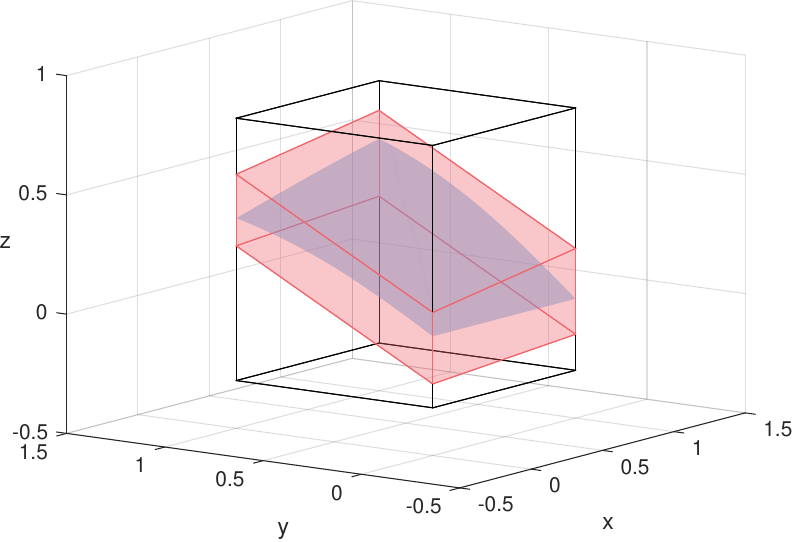}
    \caption{The truncated rectangular prism formed by four interval constraint planes and two relaxation planes.}
    \label{object-b}
\end{figure}

The problem in the distance-based method is the independent solving of the upper and lower planes without considering their relationship. One can investigate the truncated rectangular prism formed by four interval constraint planes $x=l_x$, $x=u_x$, $y=l_y$, $y=u_y$ and two linear relaxation planes $A_l \cdot x+B_l \cdot y+C_l$ and $A_u\cdot x+B_u \cdot y+C_u$, as shown in Fig.~\ref{object-b}. The curve surface is contained in the prism. Since the four side planes 
are fixed, an intuitive idea is to make the prism ``smaller'' for a tighter approximation. Obviously, the metric of volume can be used to measure the approximation of the upper/lower planes and the curve surface.  

The volume of prism can be computed as the area of the rectangular base multiplied by the height of the centroid line (see Appendix A for detailed proof). Let $z_{c}^{(u)} = A_u \cdot \frac{l_x+u_x}{2}+B_u \cdot  \frac{l_y+u_y}{2}+C_u$, and $z_{c}^{(l)} = A_l \cdot \frac{l_x+u_x}{2}+B_l \cdot  \frac{l_y+u_y}{2}+C_l$. Thus, the prism's volume $V$ in Fig.~\ref{object-b} can be calculated with
\begin{equation}
    \begin{aligned}
        \mathit{height} &=z_{c}^{(u)} - z_{c}^{(l)}, \\
        V &= (u_x - l_x) \cdot (u_y - l_y) \cdot \mathit{height}.
    \end{aligned}
    \tag{9}
\end{equation}

It is observed from Eq.~9 that the volume does not depend on the sampling points, thus reducing the influence of sampling randomness. Since $l_x$, $u_x$, $l_y$ and $u_y$ are known parameters, the volume-based method is also a linear programming problem: 
\begin{equation}
    \begin{aligned}
        & \min \limits_{A_u,B_u,C_u,A_l,B_l,C_l} V \\
        \text{s.t.} \; &A_l \cdot x + B_l \cdot y + C_l \leq f(x,y), \forall (x,y) \in B, \\ 
        &A_u \cdot x + B_u \cdot y + C_u \geq f(x,y), \forall (x,y) \in B.
    \end{aligned}
    \tag{10}
\end{equation}

The volume $V$ directly reflects the degree of spatial looseness between the upper and lower planes, and minimizing the volume naturally creates a synergistic relationship between them. Furthermore, the volume-based method requires to solve only one LP problem, while the distance-based method needs two. This can accelerate the calculation.

\subsection{Hybrid Volume-Area-based Method (\emph{DeepPrism})}
\label{hybrid}

\begin{figure}[t]
    \centering
    \includegraphics[width=0.4\textwidth]{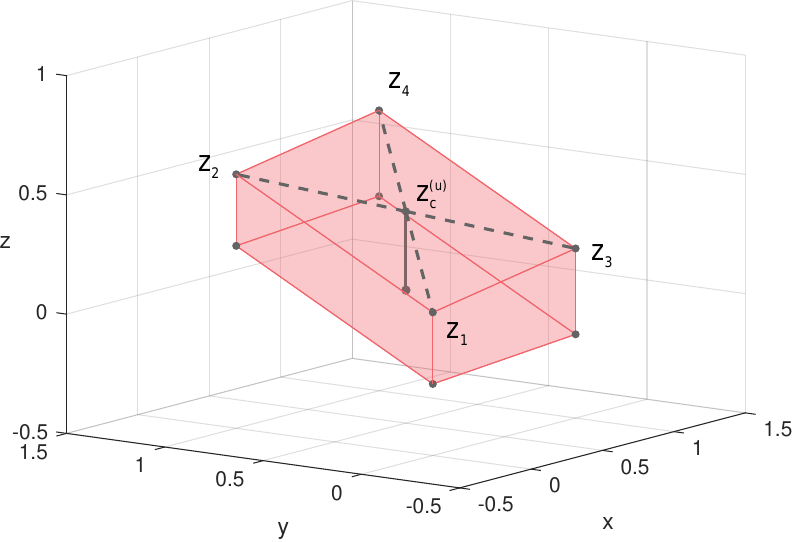}
    \caption{Two important lines of the truncated rectangular prism: The solid line connects the centroids in the upper and lower planes, and the dashed line connects $Z_u$ and $Z_l$. The solid line ``controls'' volume, and the dashed line ``controls'' surface area.}
    \label{object-c}
\end{figure}

Eq.~9 indicates that the centroid line in Fig.~\ref{object-c} controls the volume, and the plane can be considered as rotating around the centroid. Maintaining roughly the same volume, the goal is to obtain a more ``rounded'' truncated rectangular prism with a smaller surface area $S$ (see Appendix C for detailed explanation). In this way, one can obtain a tighter abstract domain in three-dimensional space. The objective function is expressed as:

\begin{equation}
    \min \limits_{A_u,B_u,C_u,A_l,B_l,C_l} \alpha \cdot V + (1-\alpha) \cdot S.
    \tag{11}
\end{equation}

$\alpha$ weights the contributions of volume and surface area. The surface area $S$ is the sum of the areas of six faces, with four trapezoidal planes on the front, back, left and right only depending on the height of the centroid line. Meanwhile, the areas of the upper and lower planes together are nonlinear, involving the calculation of squares and square roots. As proved in Appendix B, the surface area $S$ is positively correlated with the sum of the absolute differences in the \emph{z}-coordinate between each of the four corner points and the center point, so we can minimize the surface area using this sum. To unify the dimensional scale, we use the centroid height to minimize the volume. Let $\{z_i\}_{i=1}^4$
denote the \emph{z}-values of the upper plane at the four corner points, computed as $z_i=A_u \cdot x+B_u \cdot y+C_u$ where $(x,y)\in \{l_x,u_x\}\times \{l_y,u_y\}$, and $\{z_i\}_{i=5}^8$
denote the \emph{z}-values of the lower plane at the four corner points, computed as $z_i=A_l \cdot x+B_l \cdot y+C_l$ where $(x,y) \in \{l_x,u_x\}\times \{l_y,u_y\}$. Then, the objective function in Eq.~11 can be updated to:
\begin{equation}
    \begin{aligned}
        sum = \sum_{i=1}^4\left|z_i-z_c^{(u)}\right| + \sum_{i=5}^8\left|z_i-z_c^{(l)}\right|, \\
        \min \limits_{A_u,B_u,C_u,A_l,B_l,C_l} \alpha \cdot height + (1-\alpha) \cdot sum.
    \end{aligned}
    \tag{12}
\end{equation}

Note that the optimization of surface area can be applied to the distance-based method \cite{ryou2021scalable}. As more points are selected, the average distance becomes closer to the centroid height. Therefore, the distance-area and volume-area methods can be unified, where the latter is more essential and elegant to avoid the sampling limitation.

\section{Verification Process}
\label{Prover}
\begin{figure*}[ht]
    \centering
    \includegraphics[width=\textwidth]{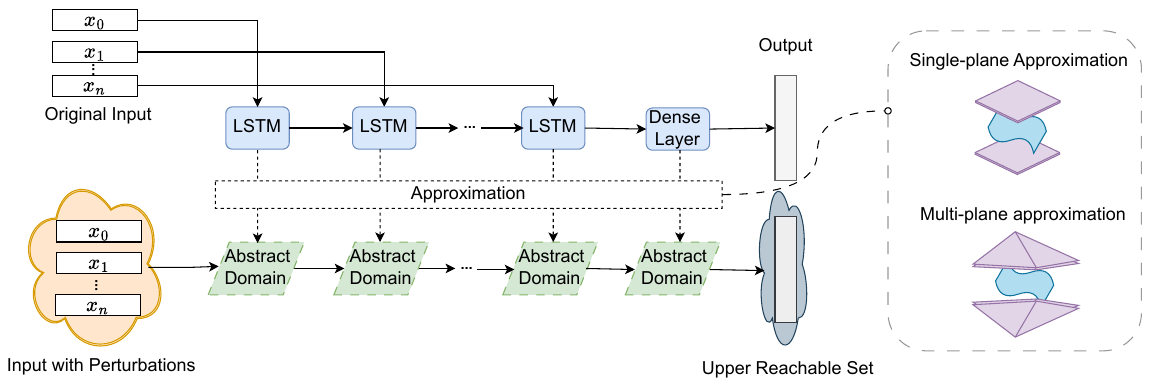}
    \caption{The process of LSTM Certification. The sequence data $X$ go through the LSTM to obtain the output, while the perturbed data $[X - \epsilon, X + \epsilon]$ are processed through abstract domains to compute the reachable set. The abstract domain is executed through either single-plane or multi-plane approximation.}
    \label{frame}
\end{figure*}

Our verification process is illustrated in Fig.~\ref{frame}, consisting of four steps: input, approximation, propagation, and output. First, we represent the original sequence data in an interval form with perturbations. Second, the perturbation propagation process is modeled with single-plane and multi-plane approximation methods as explained follows. Third, we use the backsubstitution of DeepPoly to reduce the imprecision. Finally, we obtain the over-approximation of the neural network reachable set.

\subsection{Single-plane Approximation}
\label{single}
The single-plane approximation uses an upper plane and a lower plane to perform linear relaxation of the nonlinear parts, thereby generating the corresponding abstract domain. Through the hybrid volumn-area method, we can obtain the tighter relaxation.

Given an input $x$ with perturbation $\epsilon$, we define a function $g(x, \epsilon, p)$ to represent the gap in predicted probabilities of true label $t$ and prediction class $p$. Thus, we formalize the robustness verification as 
$\forall p \neq t, g(x, \epsilon, p) \geq 0$. The single-plane approximation achieves this by solving the LP problem (Eq.~12), obtaining the coefficients of the upper and lower planes, and calculating the value of $g(x, \epsilon, p)$.

In Eq.~12, we balance the influence of volume and surface area by adjusting the weight $\alpha$. Specifically, $\alpha$ is constrained in $[0, 1]$, and an exhaustive search is performed with a step size of $0.001$. For each $\alpha$, the average value of the objective function $g(x, \epsilon, i)$ is computed, and the $\alpha$ that yields the maximum average value is considered optimal. We conduct experiments and find that the proper value of $\alpha$ is $0.674$. This value indicates that giving a slight preference to volume helps to improve the verification accuracy.

\subsection{Multi-plane Approximation}
\label{multi}
The single-plane volume-area method only produces a single bound and does not consider global properties of neural networks. Further, this method is, in a sense, greedy: Selecting locally optimal planes for each neuron does not necessarily lead to global optimization. A natural idea is a divide-and-conquer strategy: Divide the LP region and then combine the sub-regions proportionally, with the proportions solved by gradient descent. In this way, the single-plane approximation is upgraded into a multi-plane approximation. 

\begin{table*}[t]
\centering
\begin{tabular}{ccccccc}
    \hline
    Name & Task & Input Type & Samples & Classes & Source \\
    \hline
    MNIST &  Image Classification & Image & $70,000$ & $10$ & \cite{726791}  \\
    Google Speech Commands (GSC) &  Speech Recognition & Audio & $105,829$ & $35$ & \cite{warden2018speech}  \\
    Free Spoken Digit Dataset (FSDD) &  Speech Recognition & Audio & $3,000$ & $10$ & GitHub  \\
    Rotten Tomatoes Movie Review (RT) & Sentiment Analysis & Text & $43,800$ & $2$ & \cite{pang-lee-2005-seeing}  \\
    \hline
\end{tabular}
\caption{Summary of dateset.}
\label{dataset}
\end{table*}

\begin{figure}[t]
    \centering
    \includegraphics[width=0.4\textwidth]{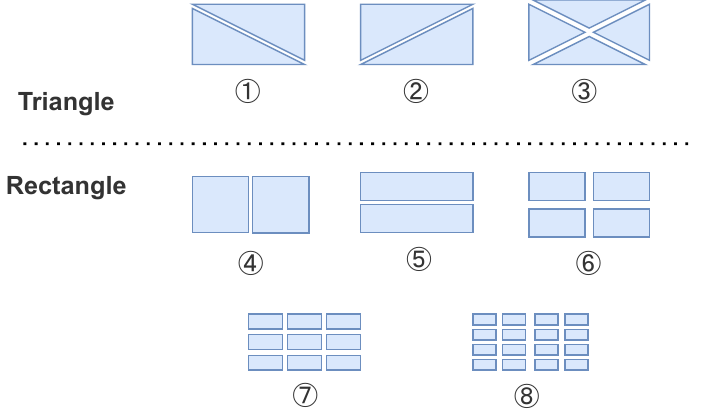}
    \caption{Divisions of multi-plane approximation, which are marked as follows: \textcircled{\raisebox{-0.9pt}{1}} 2-tri-up, \textcircled{\raisebox{-0.9pt}{2}} 2-tri-down, \textcircled{\raisebox{-0.9pt}{3}} 4-tri, \textcircled{\raisebox{-0.9pt}{4}} 2-rec-vec, \textcircled{\raisebox{-0.9pt}{5}} 2-rec-hor, \textcircled{\raisebox{-0.9pt}{6}} 4-rec, \textcircled{\raisebox{-0.9pt}{7}} 9-rec, \textcircled{\raisebox{-0.9pt}{8}} 16-rec.}
    \label{split}
\end{figure}

As shown in Fig.~\ref{split}, we divide the base of the truncated rectangular prism, $[l_x, u_x] \times [l_y, u_y]$ into different parts, such as triangular or rectangular sub-regions. Theoretically, finer divisions improve verification accuracy with higher computational cost, so it is necessary to find a trade-off between accuracy and efficiency.

The sub-regions are denoted as $\tau_k$, where $k$ is the index of the subdivided region, and $\tau_0$ represents the initial region. For each sub-region, linear programming is performed to obtain the piecewise upper and lower planes.

Let $LB^0$ represent the lower plane for the entire region, $LB^k$ represent the lower plane for the sub-region $\tau_k$, and the final plane $LB$ is expressed as a linear combination of $LB^k$:
\begin{equation}
    LB = \sum_{k=0} \lambda_k \cdot LB^k, \; \sum_{k=0} \lambda_k = 1.
    \tag{13}
\end{equation}

Therefore, the robustness verification function $g(x, \epsilon, p)$ is now associated with the weight matrix $\boldsymbol{\lambda}=\{\lambda_0,\, \lambda_1,\, \lambda_2,\,\ldots \}$ and is redefined as $g(x, \epsilon, p, \boldsymbol{\lambda})$. To find $\boldsymbol{\lambda}$, we solve the optimization problem for each prediction label $p$:
\begin{equation}
    \max_{\boldsymbol{\lambda}} g(x, \epsilon, p, \boldsymbol{\lambda}) \geq 0,
    \tag{14}
\end{equation}
which can be solved by the gradient descent algorithm in machine learning. Define the loss function as $L = -g(x, \epsilon, p, \boldsymbol{\lambda})$, and update $\boldsymbol{\lambda}$ based on this loss. $L < 0$ indicates that a $\boldsymbol{\lambda}$ has been found that ensures robustness. For specific details, see \cite{ryou2021scalable}. Using this approach, we refine the abstract domain, allowing us to tighten the output reachable set and improve the verification accuracy.

\section{Experiments}
\label{exp}
We evaluate the effectiveness of \emph{DeepPrism} on RNN robustness verification.  Specifically, we consider three research questions and answer them respectively.

RQ1: How is the verification accuracy of the single-plane \emph{DeepPrism} compared to \emph{RNN-Guard} and \emph{Prover}? 

RQ2: Can \emph{DeepPrism} further improve the verification accuracy in multi-plane approximation?

RQ3: What is the impact of different refinement divisions on the accuracy and running time?

\subsection{Experimental Setup}
\paragraph{Environment.} All experiments are executed on a Linux server with the configuration of NVIDIA GeForce 4090, Intel i9-13900K CPU, and 64GB RAM. We use PyTorch 2.4 to implement all models, Gurobi 11.0 as the LP solver.

\paragraph{Dataset.} Four datasets corresponding to three tasks are used and listed in Table~\ref{dataset}. Specifically, (1) MNIST for image classification; (2) GSC and FSDD for speech recognition; (3) RT for sentiment analysis. Appendix D provides a description of the dataset processing.

\paragraph{Parameters.} We consider three parameters of LSTM: the frame $f$, the dimension of the hidden state $h$, and the number of layers $\ell$. We use $\epsilon $ to denote the perturbation, noting that larger perturbations result in a validation accuracy close to 0, which has no practical significance.

\paragraph{Baselines.} Four abstract interpretation-based methods are introduced in the related work. Among them, \emph{Prover} achieves the highest precision and largest scale, making it the current state-of-the-art (SOTA) technique. In addition, \emph{RNN-Guard} extends the evaluation to text data and achieves promising results. So, we select \emph{Prover} and \emph{RNN-Guard} as the baselines in the experiments.

\subsection{Verification of Single-plane Approximation (RQ1)}
\subsubsection{Image Classification.}
\begin{figure}[ht]
    \centering
    \includegraphics[width=0.5\textwidth]{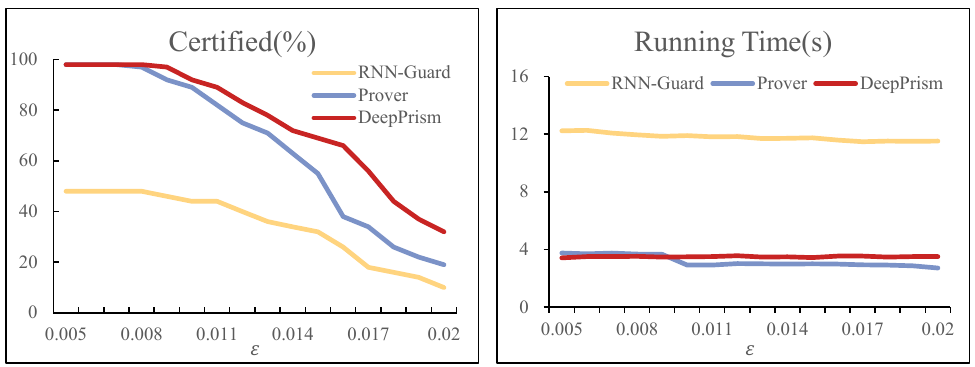}
    \caption{Results on MNIST with different perturbation and models where $f=4, h=32$ and $\ell=2$.}
    \label{res1}
\end{figure}

Fig.~\ref{res1} shows the verification comparison of three models (\emph{Prover}, \emph{RNN-Guard}, \emph{DeepPrism}) using single-plane approximation.
As the perturbation increases, the verification accuracy decreases, with the advantages of \emph{DeepPrism} becoming more evident. \emph{DeepPrism} outperforms other baselines on accuracy with a slight but acceptable increase in computation time. See Appendix E for more results under different parameter settings. 

As to the impact of the model parameters, we set a representative perturbation of $0.012$. The performance of three models at different $f$, $h$, and $\ell$ is shown in Table~\ref{res2}. Vertically, an increase in $f$ and $\ell$ leads to a decline in the accuracy of the verifier, while an increase in $h$ causes an improvement. Horizontally, \emph{DeepPrism} outperforms other baselines under all configurations.

\begin{table}[ht]
\centering
\begin{tabular}{@{}rrr|rr|rr|rr@{}}
\toprule
\multirow{2}{*}{$f$} & \multirow{2}{*}{$h$}   & \multirow{2}{*}{$\ell$}   & \multicolumn{2}{c|}{\emph{RNN-Guard}} & \multicolumn{2}{c|}{\emph{Prover}} & \multicolumn{2}{c}{\emph{DeepPrism}} \\
 & &  & Acc. & Time & Acc. & Time & Acc. & Time \\
\midrule
4 & 32 & 1 & 40 & 5.84 & 45 & 1.47 & \textbf{83} & 1.79  \\
4 & 32 & 2 & 25 & 11.83 & 49 & 3.02 & \textbf{68} & 3.57 \\
4 & 32 & 3 & 20 & 17.71 & 25 & 4.56 & \textbf{59} & 5.25 \\
4 & 64 & 1 & 53 & 11.60 & 81 & 2.84 & \textbf{88} & 3.15 \\
4 & 128 & 1 & 62 & 22.84 & 86 & 5.46 & \textbf{93} & 6.18 \\
7 & 32 & 1 & 7 & 10.89 & 14 & 2.70 & \textbf{53} & 3.03 \\
\bottomrule
\end{tabular}
\caption{Comparison of three models with different neural network on MNIST under the perturbation $\epsilon=0.012$.}
\label{res2}
\end{table}

\subsubsection{Speech Recognition.}
We compare three baseline methods on the GSC and FSDD datasets, as shown in Table~\ref{ressr1} and Table~\ref{ressr2}. \emph{DeepPrism} demonstrates superior performance by achieving the highest accuracy while simultaneously maintaining the shortest runtime, reflecting an optimal balance between effectiveness and computational efficiency.

\setlength{\tabcolsep}{3pt}
\begin{table*}[ht]
\centering
\begin{tabular}{@{}c|lr|lr|lr|lr|lr|lr|lr|lr@{}}
    \toprule
     \multicolumn{1}{c|}{ } & \multicolumn{2}{c|}{\textcircled{\raisebox{-0.9pt}{1}} 2-tri-up} & \multicolumn{2}{c|}{\textcircled{\raisebox{-0.9pt}{2}} 2-tri-down} & \multicolumn{2}{c|}{\textcircled{\raisebox{-0.9pt}{3}} 4-tri} & \multicolumn{2}{c|}{\textcircled{\raisebox{-0.9pt}{4}} 2-rec-vec} & \multicolumn{2}{c|}{\textcircled{\raisebox{-0.9pt}{5}} 2-rec-hor} & \multicolumn{2}{c|}{\textcircled{\raisebox{-0.9pt}{6}} 4-rec} & \multicolumn{2}{c|}{\textcircled{\raisebox{-0.9pt}{7}} 9-rec} & \multicolumn{2}{c}{\textcircled{\raisebox{-0.9pt}{8}} 16-rec} \\
     \midrule
    $\epsilon$ & Acc. & Time & Acc. & Time & Acc. & Time & Acc. & Time & Acc. & Time & Acc. & Time & Acc. & Time  & Acc. & Time \\
    \midrule
    0.005 & \textbf{97} & 7.79 & \textbf{97} & 7.80 & \textbf{97} & 14.58 & \textbf{97} & 10.81 & \textbf{97} & 8.42 & \textbf{97} & 18.69 & \textbf{97} & 28.18 & \textbf{97} & 38.02 \\

    0.008 & 94 & 8.82 & 94 & 9.14 & 94 & 15.97 & 93 & 11.17 & 94 & 10.93 & 95 & 18.98 & \textbf{97} & 28.40 & \textbf{97} & 37.85 \\

    0.011 & 88 & 9.84 & 87 & 9.55 & 90 & 17.56 & 81 & 12.42 & 95 & 11.55 & \textbf{97} & 19.00 & \textbf{97} & 29.89 & \textbf{97} & 37.85 \\

    0.014 & 68 & 12.13 & 72 & 13.68 & 72 & 16.87 & 54 & 19.62 & 86 & 17.26 & 93 & 19.74 & 95 & 33.09 & \textbf{96} & 38.69 \\

    0.017 & 44 & 13.08 & 39 & 16.88 & 45 & 19.86 & 23 & 19.17 & 64 & 18.98 & 65 & 21.42 & 71 & 33.15 & \textbf{76} & 37.64 \\

    0.020 & 18 & 14.47 & 18 & 21.91 & 20 & 22.07 & 7 & 17.06  & 28 & 19.53 & 54 & 21.72  & 57 & 31.68  & \textbf{64} & 38.39 \\
    \bottomrule
\end{tabular}
\caption{Verification accuracy of different divisons on MNIST under different perturbations where $f=4, h=32$ and $\ell=2$.}
\label{res5}
\end{table*}

\setlength{\tabcolsep}{3pt}
\begin{table}[ht]
\centering
    \begin{tabular}{@{}c|rr|rr|rr@{}}
        \toprule
        \multirow{2}{*}{$\epsilon$(dB)} & \multicolumn{2}{c|}{\emph{RNN-Guard}} & \multicolumn{2}{c|}{\emph{Prover}} & \multicolumn{2}{c}{\emph{DeepPrism}} \\
        \cline{2-7}
        & Acc. & Time & Acc. & Time & Acc. & Time \\
        \midrule
        -100 & 20 & 57.12 & 44 & 18.06 & \textbf{46} & 15.06 \\
        -95 & 6 & 59.27 & 28 & 17.80 & \textbf{29} & 15.14 \\
        -90 & 0 & - & \textbf{14} & 17.75 & \textbf{14} & 15.03 \\ 
        -85 & 0 & - & 6 & 17.61 & \textbf{8} & 14.97 \\
        -80 & 0 & - & \textbf{3} & 17.63 & \textbf{3} & 14.98 \\
        \bottomrule
    \end{tabular}
    \caption{Comparison of three models on GSC dataset.}
    \label{ressr1}
\end{table}

\setlength{\tabcolsep}{3pt}
\begin{table}[ht]
\centering
    \begin{tabular}{@{}c|rr|rr|rr@{}}
        \toprule
        \multirow{2}{*}{$\epsilon$(dB)} & \multicolumn{2}{c|}{\emph{RNN-Guard}} & \multicolumn{2}{c|}{\emph{Prover}} & \multicolumn{2}{c}{\emph{DeepPrism}} \\
        \cline{2-7}
        & Acc. & Time & Acc. & Time & Acc. & Time \\
        \midrule
        -100 & 51 & 68.32 & \textbf{97} & 21.96 & \textbf{97} & 20.99 \\
        -95 & 39 & 90.35 & \textbf{90} & 21.93 & \textbf{90} & 21.17 \\
        -90 & 27 & 94.81 & 86 & 21.74 & \textbf{88} & 20.89 \\ 
        -85 & 11 & 111.98 & 81 & 21.80 & \textbf{84} & 20.89 \\
        -80 & 0 & - & 70 & 21.78 & \textbf{72} & 21.00 \\
        \bottomrule
    \end{tabular}
    \caption{Comparsion of  three models on FSDD dataset.}
    \label{ressr2}
\end{table}

\subsubsection{Sentiment Analysis.}
We compare three baseline methods on the RT dataset, as shown in Table~\ref{ressa}. \emph{DeepPrism} achieves the highest accuracy, indicating its effectiveness in sentiment classification.

\setlength{\tabcolsep}{3pt}
\begin{table}[ht]
\centering
    \begin{tabular}{@{}c|rr|rr|rr@{}}
        \toprule
        \multirow{2}{*}{$\epsilon$} & \multicolumn{2}{c|}{\emph{RNN-Guard}} & \multicolumn{2}{c|}{\emph{Prover}} & \multicolumn{2}{c}{\emph{DeepPrism}} \\
        \cline{2-7} 
        & Acc. & Time & Acc. & Time & Acc. & Time \\
        \midrule
        0.05 & 51 & 57.12 & 77 & 22.15 & \textbf{84} & 22.84\\
        0.07 & 39 & 58.85 & 48 & 23.31 & \textbf{59} & 23.04\\
        0.09 & 27 & 59.27 & 27 & 25.32 & \textbf{28} & 24.87\\ 
        0.11 & 11 & 58.98 & 11 & 22.46 & \textbf{16} & 22.74\\
        0.13 & 0 & - & 10 & 25.43 & \textbf{12} & 22.08\\
        0.15 & 0 & - & 0 & 26.32 & \textbf{2} & 23.90\\ 
        \bottomrule
    \end{tabular}
    \caption{Comparison of three models on RT dataset.}
    \label{ressa}
\end{table}

\subsection{Verification of Multi-plane Approximation (RQ2)}
Theoretically, the results of multi-plane approximation should be better than those of single-plane approximation because using more planes provides a closer approximation to the curve surface. The experimental results in Fig.~\ref{res3} confirm this (See Appendix F for more results). 

At the same time, we observe that the impact of different approximations on the multi-plane approximation method is not significant. \emph{DeepPrism} performs slightly better but with more time consumption, followed by \emph{Prover}. This is because the division process and gradient descent dilute the impact of linear programming, achieving similar results with more iterations.

\begin{figure}[ht]
    \centering
    \includegraphics[width=0.5\textwidth]{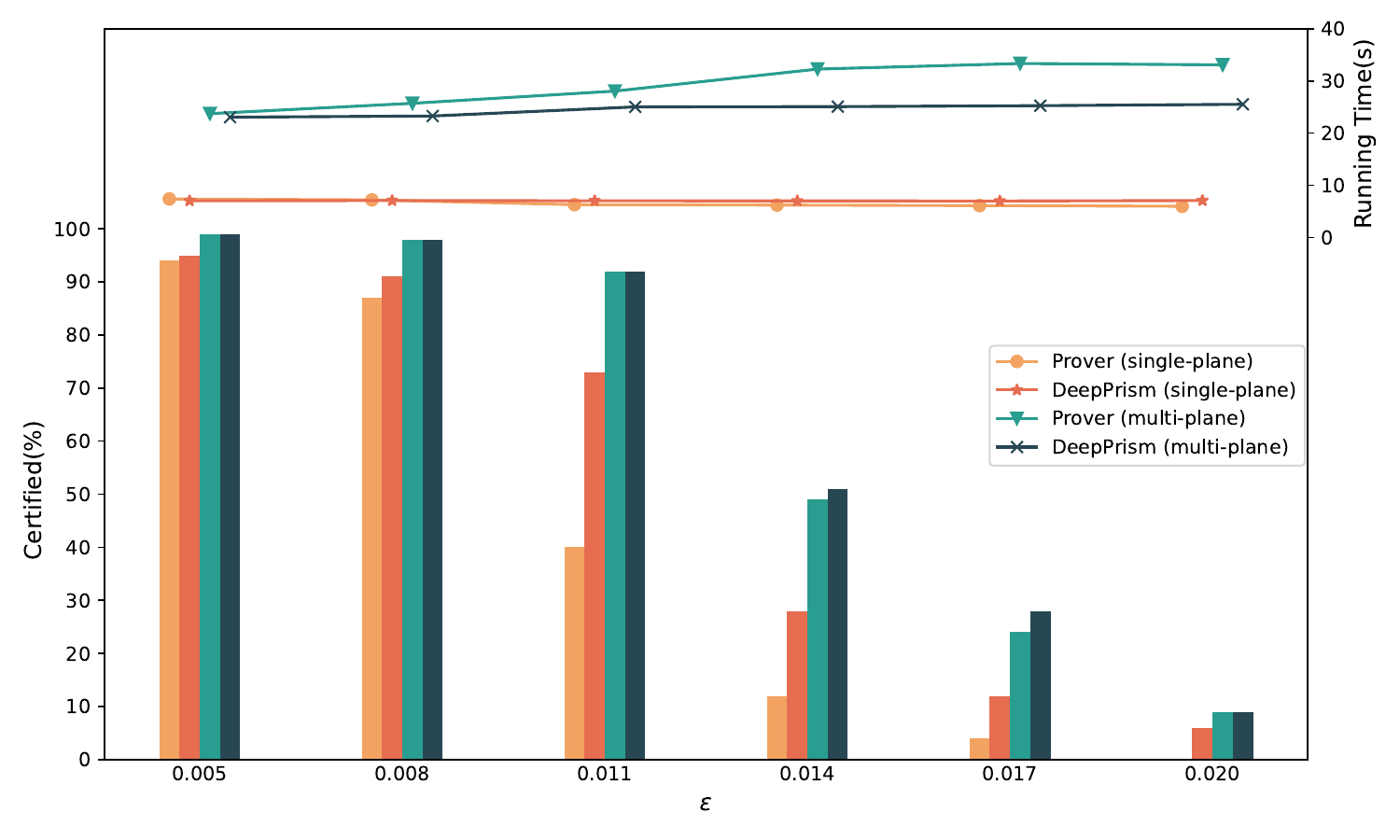}
    \caption{Comparison of two verification methods (single-plane, multi-plane) on two models (\emph{Prover} and \emph{DeepPrism}) evaluated on MNIST using the LSTM model with $f=4$, $h=32$ and $\ell=3$.}
    \label{res3}
\end{figure}

\subsection{Effect of Refinement Divisions (RQ3)}
Finally, we investigate the impact of different divisions on multi-plane approximation. Table~\ref{res5} presents the experimental results under different perturbations (See Appendix G for more results). For \textcircled{\raisebox{-0.9pt}{1}} to \textcircled{\raisebox{-0.9pt}{6}}, under low perturbations, all models perform similarly under all divisions, where division \textcircled{\raisebox{-0.9pt}{3}} 4-tri and \textcircled{\raisebox{-0.9pt}{6}} 4-rec slightly outperform others. Under high perturbations, division \textcircled{\raisebox{-0.9pt}{6}} 4-rec clearly outperforms others, maintaining higher accuracy. One can observe that rectangular division provides more uniform coverage of the region, reducing the boundary effects. 

In addition, we also test more finer divisions, with the results shown in Division \textcircled{\raisebox{-0.9pt}{4}}, \textcircled{\raisebox{-0.9pt}{6}}, \textcircled{\raisebox{-0.9pt}{7}} and \textcircled{\raisebox{-0.9pt}{8}}. Finer divisions can better capture surface features, thereby improving overall approximation performance. In the cases of \textcircled{\raisebox{-0.9pt}{7}} 9-rec and \textcircled{\raisebox{-0.9pt}{8}} 16-rec divisions, the verification accuracy has a significant improvement compared to that of \textcircled{\raisebox{-0.9pt}{6}} 4-rec division. Note that more divisions may have a computational burden. 

\section{Conclusion}
\label{conclusion}
Neural network verification plays a crucial role in ensuring the robustness of AI models. We introduce a novel over-approximation method based on the truncated rectangular prism, supported by theoretical guarantees.  This method can be effectively applied to RNN verification within the framework of abstract interpretation, leading to significant improvements in experimental results. This work not only improves existing robustness verification techniques but also offers fresh insights into nonlinear analysis in abstract interpretation. Future work will focus on reducing computational overhead and enhancing the time efficiency of the approach.

\section*{Acknowledgements}
This work is supported by the National Key Research Project of China (No. 2023YFA1009402), NSFC (Nos. 62272416, 62372176), and Huawei.

\bibliography{aaai2026}

\clearpage
\appendix

The appendix is organized as follows. Appendix A and B provides the proof of volume and surface area for truncated poly-prism, respectively. Appendix C demonstrates how different over-approximations affect verification results, and Appendix D describes the details of the experimental implementation. Appendix E, F, and G present the detailed results for RQ1, RQ2 and RQ3, respectively.
\section{Appendix A: Proof of Volume for Truncated Poly-Prism}
\noindent First, we present the definition of the truncated poly-prism. 
\begin{definition}
As shown in Figure~\ref{fig4}, let $P_1P_2\cdots P_n$ be a convex polygon in the x-y plane of $\mathbb{R}^3$, where the cooridinate of $P_i\in \mathbb{R}^3$ is $(x_i,y_i,0)$. Let $Q_i=(x_i,y_i,z_i)\in \mathbb{R}^3\,(i=1,2,\ldots,n)$ satisfies $z_1,z_2,\ldots,z_n>0$ and $Q_1,Q_2,\ldots,Q_n$ in the same plane. The geometry formed by $P_1P_2\cdots P_n; Q_1Q_2\cdots Q_n$ is referred to as ``truncated poly-prism'', and the polygon $P_1P_2\cdots P_n$ is referred to as the base of the truncated poly-prism.
\end{definition}
\setcounter{figure}{0}
\begin{figure}[ht]
    \centering
    \includegraphics[width=0.5\linewidth]{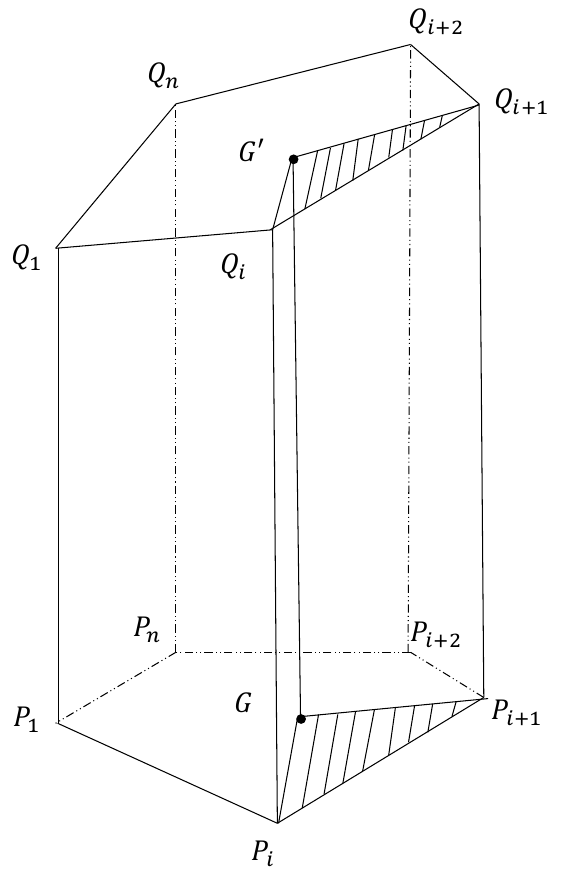}
    \caption{A truncated poly-prism.}
    \label{fig4}
\end{figure}

In this appendix, we will prove the following Theorem~1.
\begin{theorem}
The volume of the truncated poly-prism in Definition 1 is given by:
$$V=\frac{1}{n}(z_1+z_2+\cdots+z_n)\cdot \text{Area}(P_1P_2\cdots P_n).$$
\end{theorem}

\begin{figure}[ht]
    \centering
    \includegraphics[width=0.45\textwidth]{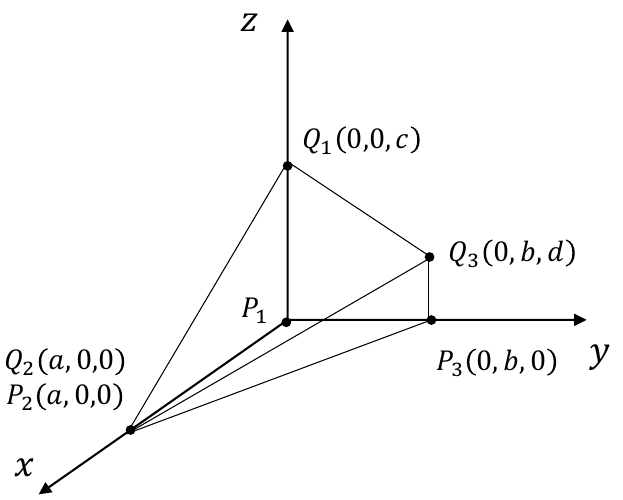}
    \caption{A truncated triangular prism, where $P_1P_2P_3$ is a right triangle.}
    \label{fig1}
\end{figure}

We present some lemmas of special cases, and then provide the whole proof of Theorem 1. 
\begin{lemma}
As shown in Figure~\ref{fig1}, let $P_1P_2P_3 \subset \mathbb{R}^2\times \{0\}$ be a right triangle, with coordinates given by:
$$
P_1=(0,0,0), P_2=(a,0,0), P_3=(0,b,0), 
$$
where $a,b>0$. Let the cooridinates of $Q_1,Q_2,Q_3\in \mathbb{R}^3$ be:
$$
Q_1=(0,0,c), Q_2=(a,0,0), Q_3=(0,b,d),
$$
where $c>0,d\geq 0$. Then the volume of the truncated triangular prism $P_1P_2P_3-Q_1Q_2Q_3$ is:
$$
V=\frac{1}{3} \cdot (c+d) \cdot \text{Area}(P_1P_2P_3). 
$$
\end{lemma}

\noindent \textbf{Proof}: The truncated triangular prism $P_1P_2P_3-Q_1Q_2Q_3$ can be seen as a quadrilateral pyramid, with the base being the trapezoid $P_1P_3Q_3Q_1$ and the height being $P_1P_2$. Therefore, its volume is:
\begin{align*}
V &= \frac{1}{3} Area(P_1P_3Q_3Q_1) \cdot |P_1P_2| \\
  &= \frac{1}{3} \left(\frac{1}{2} (c+d) \cdot b \right) \cdot a \\
  &= \frac{1}{6} (c+d)ab  \\
  &= Area(P_1P_2P_3) \cdot \frac{1}{3}(c+d).
\end{align*}

Q.E.D.\qed
\\

Lemma~1 can be extended to a more general case, where the triangle $P_1P_2P_3$ is not necessarily a right triangle. We describe this case in Lemma~2.

\begin{figure}[ht]
    \centering
    \includegraphics[width=0.45\textwidth]{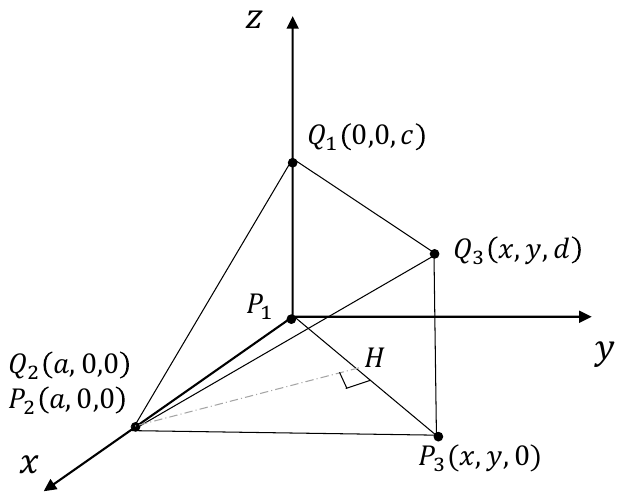}
    \caption{A truncated triangular prism, where $P_1P_2P_3$ is not a right triangle.}
    \label{fig2}
\end{figure}

\begin{lemma}
As shown in Figure~\ref{fig2}, let $P_1P_2P_3\subset \mathbb{R}^2\times \{0\}$ be a general triangle, with coordinates given by:
$$
P_1=(0,0,0), P_2=(a,0,0), P_3=(x,y,0), 
$$
where $a,x,y>0$. Let the coordinates of $Q_1,Q_2,Q_3\in \mathbb{R}^3$ be:
$$
Q_1=(0,0,c), Q_2=(a,0,0), Q_3=(x,y,d),
$$
where $c>0,d\geq 0$. Then, the volume of $P_1P_2P_3-Q_1Q_2Q_3$ is: 
$$
V=\frac{1}{3} \cdot (c+d) \cdot Area(P_1P_2P_3).
$$
\end{lemma}

\noindent \textbf{Proof}: Draw a perpendicular from $P_2$ to $P_1P_3$, and let the foot of the perpendicular be $H$. Then the truncated triangular prism can be seen as a quadrilateral pyramid with a trapezoidal base $P_1P_3Q_3Q_1$ and a height $P_2H$. Therefore, its volume is:

\begin{align*}
    V &=\frac{1}{3} Area(P_1P_3Q_3Q_1)\cdot P_2H \\
    &=\frac{1}{3} \cdot \left(\frac{1}{2}(c+d)\cdot P_1P_3\right)\cdot P_2H \\
    &=\frac{1}{3} (c+d) \cdot \frac{1}{2}{P_1P_3}\cdot P_2H \\
    &=\frac{1}{3}(c+d) \cdot Area(P_1P_2P_3).
\end{align*}

Q.E.D.\qed
\\

We now prove a more general case as follows.

\begin{figure}[ht]
    \centering
    \includegraphics[width=0.45\textwidth]{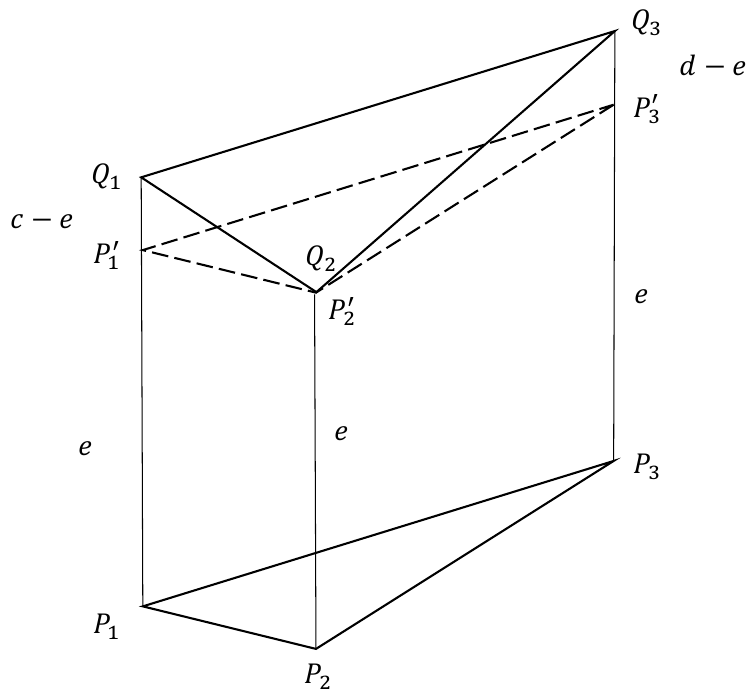}
    \caption{A truncated triangular prism.}
    \label{fig3}
\end{figure}

\begin{lemma}
As shown in Figure~\ref{fig3}, let $P_1P_2P_3\subset \mathbb{R}^2\times \{0\}$ be a general triangle, with coordinates given by:
$$
P_1=(0,0,0), P_2=(a,0,0), P_3=(x,y,0), 
$$
where $a,x,y>0$. Let the coordinates of $Q_1,Q_2,Q_3\in \mathbb{R}^3$ be:
$$
Q_1=(0,0,c), Q_2=(a,0,e), Q_3=(x,y,d),
$$
where $c>0,d,e\geq 0$. Then the volume of $P_1P_2P_3-Q_1Q_2Q_3$ is:
$$
V=\frac{1}{3} \cdot (c+d+e)\cdot Area(P_1P_2P_3).
$$
\end{lemma}

\noindent \textbf{Proof}: Without loss of generality, assume $e\leq d$. By slicing the truncated triangular prism $P_1P_2P_3-Q_1Q_2Q_3$ with the plane $z=e$, we obtain two geometric bodies: the triangular prism $P_1P_2P_3-P^{\prime}_1P^{\prime}_2P^{\prime}_3$ below the plane $z=e$ and  the truncated triangular prism $P^{\prime}_1P^{\prime}_2P^{\prime}_3-Q_1Q_2Q_3$ above the plane $z=e$, with the cross-section being the triangle $P^{\prime}_1P^{\prime}_2P^{\prime}_3$. The volumes of these two geometric bodies are:

\begin{align*}
    V_1&=Area(P_1P_2P_3)\cdot e, \\
    V_2&=\frac{1}{3} \cdot Area(P^{\prime}_1P^{\prime}_2P^{\prime}_3) \cdot ((c-e)+0+(d-e)) \\
    &=\frac{1}{3} \cdot Area(P_1P_2P_3) \cdot(d+c-2e).
\end{align*}

So, the volume of $P_1P_2P_3-Q_1Q_2Q_3$ is:
\begin{align*}
    V&=V_1+V_2   \\
    &=Area(P_1P_2P_3) \cdot e +\frac{1}{3}Area(P_1P_2P_3) \cdot (d+c-2e) \\
    &=\frac{1}{3}Area(P_1P_2P_3) \cdot (d+e+c).
\end{align*}

Q.E.D.\qed
\\

Finally, we present the whole proof of Theorem 1.

\noindent \textbf{Proof (Theorem 1)}: As shown in Figure~\ref{fig4}, let $G$ be the centroid of the convex polygon $P_1P_2\cdots P_n$, and $G^{\prime}$ be the centroid of $Q_1Q_2\cdots Q_n$. Let $z_0=\frac{1}{n}\sum_{i=1}^{n} z_i$. The coordinates of $G$ and $G^{\prime}$ are:

\begin{align*}
    G&=\left(\frac{1}{n}\sum_{i=1}^{n} x_i, \frac{1}{n}\sum_{i=1}^{n} y_i, 0\right), \\
    G^{\prime}&=\left(\frac{1}{n}\sum_{i=1}^{n} x_i, \frac{1}{n}\sum_{i=1}^{n} y_i, z_0\right).
\end{align*}

The truncated poly-prism $P_1P_2\cdots P_n-Q_1Q_2\cdots Q_n$ can be partitioned into the union of the following $n$ pairwise disjoint truncated triangular prisms:
$$
P_{i}P_{i+1}G-Q_{i}Q_{i+1}G^{\prime}, \quad
i=1,2,\ldots,n.
$$
where $P_{n+1}=P_1$ and $Q_{n+1}=Q_1$. The volumes of these $n$ truncated triangular prisms are:
$$
V_i=\frac{1}{3} Area(P_{i}P_{i+1}G)\cdot (z_i+z_{i+1}+z_{0}), \; i=1,2,\ldots,n.
$$
The volume of the polyhedron $P_1P_2\cdots P_n-Q_1Q_2\cdots Q_n$ is the sum of the volumes of the $n$ truncated triangular prisms, that is:
\begin{align*}
    V&=V_1+V_2+\cdots+V_n   \\
    &=\frac{1}{3}\sum_{i=1}^{n}\left( Area(P_iP_{i+1}G)\cdot (z_i+z_{i+1}+z_0)\right).
\end{align*}

Since $G$ is the centroid of the convex polygon $P_1P_2\cdots P_n$, it follows that:
$$
\text{Area}(P_{i}P_{i+1}G)=\frac{1}{n}\cdot \text{Area}(P_1P_2\cdots P_n), 
$$
where $i=1,2,\ldots,n$. Therefore,
$$
V=\frac{1}{3}\cdot \frac{1}{n} Area(P_1P_2\cdots P_n)\cdot \sum_{i=1}^{n}\left(z_i+z_{i+1}+z_0\right).
$$

It is easy to obtain that:
$$
\sum_{i=1}{n}\left(z_i+z_{i+1}+z_0\right)=2\sum_{i=1}^{n} z_i+n\cdot z_0
=3\sum_{i=1}^{n} z_i.
$$

Thus, we finally obtain the volume of the truncated poly-prism:
\begin{align*}
    V&=\frac{1}{n}\cdot Area(P_1P_2\cdots P_n)\cdot \sum_{i=1}^{n} z_i \\ 
    &=\frac{1}{n}(z_1+z_2+\cdots+z_n)\cdot Area(P_1P_2\cdots P_n).
\end{align*}

Q.E.D.\qed

Note that in Theorem 1, when $P_1P_2\cdots P_n$ is a rectangle, it represents the special case needed in the paper.

\section{Appendix B: Proof of Surface Area for Truncated Rectangle Prism}
\begin{figure}[ht]
    \centering
    \includegraphics[width=0.45\textwidth]{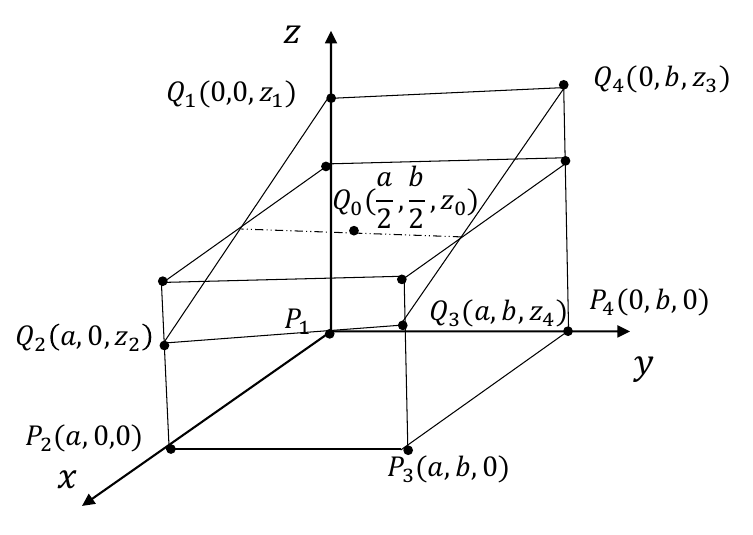}
    \caption{A truncated rectangle prism.}
    \label{fig5}
\end{figure}

Given a truncated rectangle prism $P_1P_2P_3P_4-Q_1Q_2Q_3Q_4$, shown in Figure~\ref{fig5}, 
let $z_0$ be the centroid and $z_1,z_2,z_3,z_4$ be the four lateral edges of the prism, where $z_1 \geq z_2, z_3$ and $z4 \leq z_2,z_3$. Assume the surface area of the prism is $S$. We will prove Theorem~2.

\begin{theorem}
    $S$ is positively correlated with $|z_1-z_0|+|z_2-z_0|+|z_3-z_0|+|z_4-z_0|$.
\end{theorem}

We first present some lemmas and then provide the whole proof of Theorem~2. 
\begin{lemma}
    As shown in Figure~\ref{fig5}, the four lateral edges satisfy $$z_1+z_4=z_2+z_3.$$
\end{lemma}

\noindent \textbf{Proof}: Let $Q_1Q_2Q_3Q_4 \subset \mathbb{R}^3$ be the top face of the truncated rectangle prism. The coordinates of this four vertices are:
$$
Q_1=(0,0,z_1),Q_2=(a,0,z_2),
$$
$$
Q_3=(a,b,z_4),Q_4=(0,b,z_3),
$$
where $a,b>0$.
Since $Q_1,Q_2,Q_3,Q_4$ are coplanar, they satisfy the following determinant:
\[
\left| \begin{matrix} 
0 & 0 & z_1 & 1 \\
a & 0 & z_2 & 1 \\
a & b & z_4 & 1 \\
0 & b & z_3 & 1
\end{matrix} \right| = 0,
\]
\noindent which can be simplified with 
\begin{equation*}
    \resizebox{.95\linewidth}{!}{$
z_1 \left| \begin{matrix} 
a & 0 & 1 \\
a & b & 1 \\
0 & b & 1
\end{matrix} \right|
-
z_2 \left| \begin{matrix} 
0 & 0 & 1 \\
a & b & 1 \\
0 & b & 1
\end{matrix} \right|
+
z_4 \left| \begin{matrix} 
0 & 0 & 1 \\
a & 0 & 1 \\
0 & b & 1
\end{matrix} \right|
-
z_3 \left| \begin{matrix} 
0 & 0 & 1 \\
a & 0 & 1 \\
a & b & 1
\end{matrix} \right|
=0
$},
\end{equation*}

$$
z_1 \cdot (a(b-b)+(ab-1))-z_2 \cdot ab+z_4 \cdot ab-z_3 \cdot ab=0,
$$

$$z_1-z_2+z_4-z_3=0,$$

i.e.$$z_1+z_4=z_2+z_3.$$

Q.E.D.\qed
\\

Next, we will prove Lemma 5.
\begin{lemma}
    Let the line connecting the centroids of the top and bottom faces be $z_0$, and the four lateral edges satisfy:
    $$z_1+z_2+z_3+z_4=4 \cdot z_0$$
\end{lemma}

\noindent\textbf{Proof}: Let $G$ be the centroid of the quadrilateral $P_1P_2P_3P_4$, and $G^{\prime}$ be the centroid of $Q_1Q_2Q_3Q_4$. The coordinates of $G$ and $G^{\prime}$ are:
\begin{align*}
    G&=\left(\frac{1}{4}\sum_{i=1}^{4} x_i, \frac{1}{4}\sum_{i=1}^{4} y_i, 0\right),\\
    G^{\prime}&=\left(\frac{1}{4}\sum_{i=1}^{4} x_i, \frac{1}{4}\sum_{i=1}^{4} y_i, \frac{1}{4}\sum_{i=1}^{4} z_i\right),
\end{align*}

and the length of $GG^{\prime}$ is:
$$z_0 = \frac{1}{4}\sum_{i=1}^{4} z_i,$$
i.e.
$$z_1+z_2+z_3+z_4=4 \cdot z_0.$$

Q.E.D.\qed
\\

Finally, we present the whole proof of Theorem 2. 

\noindent \textbf{Proof (Theorem 2)}: The surface area of the truncated rectangular prism is the sum of the areas of six faces, denoted as $S=\sum_{i=1}^{6} S_i$. Among these, the four lateral faces are trapezoidal, and their areas can be calculated as:
\begin{align*}
    &S_1+S_2+S_3+S_4= \frac{1}{2} \cdot (z_1+z_2) \cdot a + \frac{1}{2} \cdot (z_2+z_3) \cdot b \\
    & + \frac{1}{2} \cdot (z_3+z_4) \cdot a +\frac{1}{2} \cdot (z_4+z_1) \cdot b \\
    &= \frac{1}{2}(z_1+z_2+z_3+z_4) \cdot a + \frac{1}{2}(z_1+z_2+z_3+z_4) \cdot b \\
    & = \frac{1}{2}(z_1+z_2+z_3+z_4) \cdot (a +b) \\
    &= 2 \cdot z_0 \cdot (a +b).
\end{align*}

\begin{figure}
    \centering
    \includegraphics[width=0.5\linewidth]{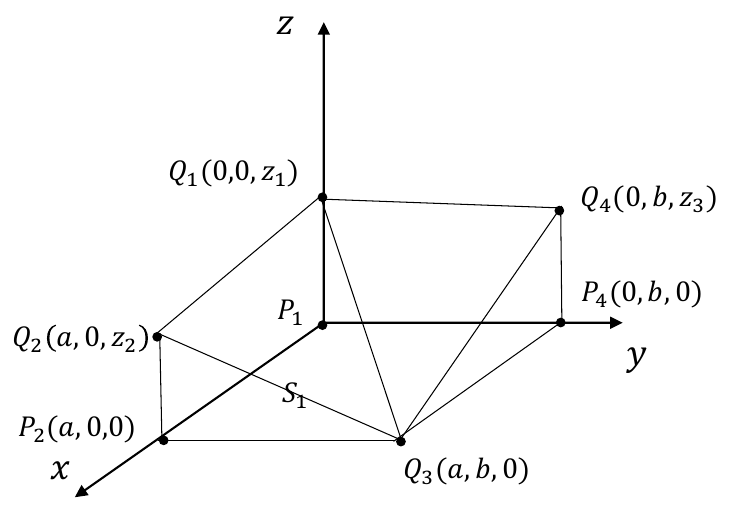}
    \caption{A truncated rectangle prism with the degeneration of $z_4=0$.}
    \label{fig6}
\end{figure}

Therefore, the four lateral faces can be seen as constants, and we only focus on the areas of the top and bottom faces. Next, let the smallest $z_4=0$, and calculate the area of the top quadrilateral, as shown in Figure \ref{fig6}.

The coordinates of the top face $Q_1Q_2Q_3Q_4 \subset \mathbb{R}^3$ are given by:
$$
Q_1=(0,0,z_1),Q_2=(a,0,z_2),
$$
$$
Q_3=(a,b,0),Q_4=(0,b,z_3),
$$
where $a,b>0$. According to Lemma 4, we have $z_1=z_2+z_3=2z_0$.

We can assume $z_2 \geq z_3$ without loss of generality. So,
\begin{align*}
&|z_1-z_0|+|z_2-z_0|+|z_3-z_0|+|0-z_0| \\
=&|z_1-\frac{z_1}{2}|+|z_2-\frac{z_1}{2}|+|z_3-\frac{z_1}{2}|+|0-\frac{z_1}{2}|\\
=& \frac{z_1}{2} + z_2-\frac{z_1}{2} + \frac{z_1}{2} - z_3 + \frac{z_1}{2}\\
=& z_1-z_3+z_2 \\
=& 2z_2.
\end{align*}

Let $a, b, c$ be three sides of a triganle, and $\Delta$ be the area of the triangle. It is well known that a highly symmetrical form of Heron's formula is:
\[
(4\Delta)^2 = \begin{bmatrix}
a^2 & b^2 & c^2
\end{bmatrix}
\begin{bmatrix}
-1 & 1 & 1 \\
1 & -1 & 1 \\
1 & 1 & -1
\end{bmatrix}
\begin{bmatrix}
a^2 \\
b^2 \\
c^2
\end{bmatrix}.
\]
The simplified expression is:
$$16\Delta^2 = (a^2+b^2+c^2)^2-2(a^4+b^4+c^4),$$
and for the triangle $Q_1Q_2Q_3$:
\begin{align*}
    16\Delta^2 =& ((z_1-z_2)^2+a^2+b^2+z_2^2+z_1^2+a^2+b^2)^2 \\
    &-2(((z_1-z_2)^2+a^2)^2+(b^2+z_2^2)^2\\
    &\quad\quad+(z_1^2+a^2+b^2)^2) \\
    =&((z_1-z_2)^2+a^2+b^2+z_2^2+z_1^2+a^2+b^2)^2 \\
    &-2((z_1-z_2)^2+2a^2(z_1-z_2)^2+a^4) \\
    &-2(z_2^4+2b^2z_2^2+b^4) \\
    &-2(z_1^4+2(a^2+b^2)z_1^2+(a^2+b^2)^2).
\end{align*}

The constant term is:
\begin{align*}
    & 4(a^2+b^2)^2-2(a^4+b^4+(a^2+b^2)^2) \\
    =& 2(a^2+b^2)^2-2(a^4+b^4) \\
    =& 4a^2b^2.
\end{align*}

The second-order term is:
\begin{align*}
    & 4(a^2+b^2)(z_1^2+z_2^2+(z_1-z_2)^2) \\
    & -4a^2(z_1-z_2)^2-4b^2z_2^2-4(a^2+b^2)z_1^2 \\
    =& 4a^2z_2^2+4b^2(z_1-z_2)^2.
\end{align*}

The four-order term is:
\begin{align*}
    & z_1^4+z_2^4+(z_1-z_2)^4 \\
    & +2z_1^2z_2^2+2z_1^2(z_1-z_2)^2+2z_2^2(z_1-z_2)^2 \\
    & -2z_1^4-2z_2^4-2(z_1-z_2)^4 \\
    =& 2z_1^2z_2^2 +2(z_1^2+z_2^2)(z_1-z_2)^2 \\
    & -z_1^4-z_2^4-(z_1-z_2)^4 \\
    =& -(z_1^2-z_2^2)^2+(z_1-z_2)^2(2z_1^2+2z_2^2-(z_1-z_2)^2) \\
    =& -(z_1-z_2)^2(z_1+z_2)^2+(z_1-z_2)^2(2z_1^2+2z_2^2-(z_1-z_2)^2)\\
    =&(z_1-z_2)^2(-(z_1+z_2)^2+2z_1^2+2z_2^2-(z_1-z_2)^2) \\
    =&0.
\end{align*}

So,
$$\Delta = \frac{1}{4}\sqrt{4a^2b^2+4a^2z_2^2+4b^2(z_1-z_2)^2}.$$

The area of the triangle $Q_1Q_3Q_4$ can be calculated in the same way. Since $z_1=z_2+z_3$, we have:
\begin{align*}
    S_{Q_1Q_2Q_3Q_4} =& S_{\triangle Q_1Q_2Q_3} + S_{\triangle Q_1Q_3Q_4} \\
    =& \frac{1}{4}\sqrt{4a^2b^2+4a^2z_2^2+4b^2(z_1-z_2)^2} \\
    &+ \frac{1}{4}\sqrt{4a^2b^2+4a^2z_3^2+4b^2(z_1-z_3)^2} \\
    =& \frac{1}{4}\sqrt{4a^2b^2+4a^2z_2^2+4b^2z_3^2} \\
    &+ \frac{1}{4}\sqrt{4a^2b^2+4a^2z_3^2+4b^2z_2^2} \\
    =& \frac{1}{2}(\sqrt{a^2b^2+a^2z_2^2+b^2z_3^2} \\
    &+\sqrt{a^2b^2+a^2z_3^2+b^2z_2^2}),
\end{align*}

There is an inequality:
\begin{align*}
    (x+y+z)^2=&x^2+y^2+z^2+2xy+2yz+2xz \\
    \leq & x^2+y^2+z^2+x^2+y^2+y^2+z^2+x^2+z^2 \\
    =&  3(x^2+y^2+z^2),
\end{align*}
i.e.
$$\sqrt{x^2+y^2+z^2} \geq \frac{1}{\sqrt{3}}(x+y+z).$$

Therefore, we have:
\begin{align*}
    &S_{Q_1Q_2Q_3Q_4} \\
    &= \frac{1}{2}(\sqrt{a^2b^2+a^2z_2^2+b^2z_3^2} +\sqrt{a^2b^2+a^2z_3^2+b^2z_2^2})\\
    &\geq \frac{1}{2}(\sqrt{a^2b^2+a^2z_2^2} +\sqrt{a^2b^2+b^2z_2^2}) \\
    &\geq \frac{1}{2\sqrt{3}}(ab+az_2+ab+bz_2) \\
    &\geq \frac{1}{\sqrt{3}}ab + \frac{a+b}{2\sqrt{3}}z_2.
\end{align*}


Given that $z_2 \geq z_3$, we have 
\begin{align*}
    &S_{Q_1Q_2Q_3Q_4} \\
    &= \frac{1}{2}(\sqrt{a^2b^2+a^2z_2^2+b^2z_3^2} +\sqrt{a^2b^2+a^2z_3^2+b^2z_2^2})\\
    &\leq \frac{1}{2}(\sqrt{a^2b^2+a^2z_2^2+b^2z_2^2} +\sqrt{a^2b^2+a^2z_2^2+b^2z_2^2}) \\
    &\leq \sqrt{a^2b^2+(a^2+b^2)z_2^2}) \\
    &\leq ab + \sqrt{a^2+b^2}z_2.
\end{align*}

Since the process of calculating the top area is the same as that for the bottom area, the surface area $S$ satisfies the expression:
$$kz_1+b \leq S \leq k^{\prime}z_1+b^{\prime}.$$
So, $S$ is positively correlated with $z_2$ as well as $|z_1-z_0|+|z_2-z_0|+|z_3-z_0|+|z_4-z_0|$.

Q.E.D.\qed
\\

\section{Appendix C: An Illustration of Approximation}
\begin{figure}[ht]
    \centering
    \includegraphics[width=0.35\textwidth]{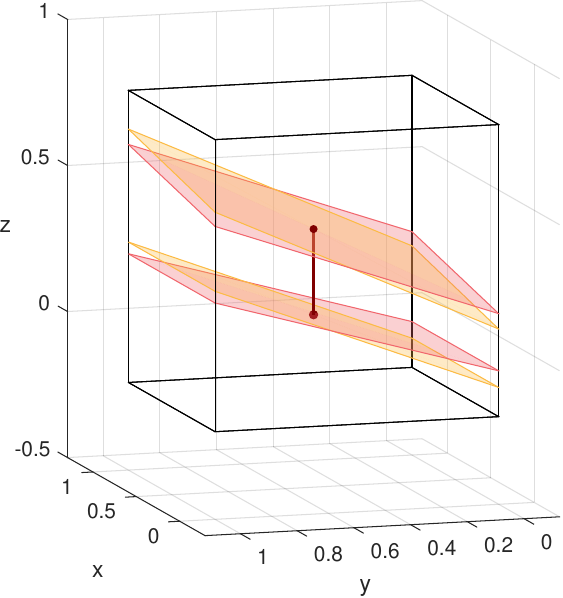}
    \caption{The two truncated rectangular prisms formed by the red and yellow planes have the same volumes but different surface areas. The red prism has a smaller surface area.}
    \label{s4}
\end{figure}

A more detailed explanation of the volume-area-based method in tighter over-approximation is provided here. As shown in Figure~\ref{s4}, the truncated rectangular prisms formed by the red and yellow planes have the same volume, since they share the same centroid line. However, their surface areas differ, with the red one having a smaller surface area. 

\begin{figure}[ht]
    \centering
    \includegraphics[width=0.35\textwidth]{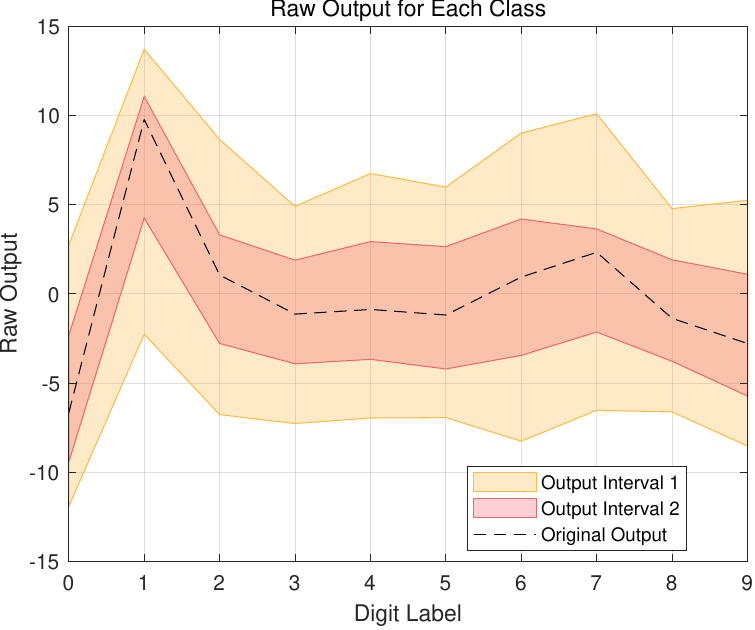}
    \caption{Different output intervals caused by two relaxations. The red prism has a smaller surface area, so the red interval is tighter than the yellow one and closer to the original output.}
    \label{s5}
\end{figure}

Different over-approximations lead to different abstract domains, which ultimately affect the output intervals, as shown in Figure~\ref{s5}. The red region is smaller than the yellow one, indicating a tighter approximation. Moreover, the lower bound of class 1 is higher than the upper bounds of all other classes, making verification successful. In contrast, the yellow region does not satisfy this condition, leading to verification failure.

\section*{Appendix D: Experimental Implementation}
\paragraph{Evaluation details of the image classification task.} In the experiments, we randomly select 100 samples from the test set, check whether these samples are correctly classified, and skip the misclassified ones. Then, we apply perturbations to the originally correctly classified sample and verify whether it remains correctly classified, i.e., the classification is robust. When the verification of a sample exceeds 120 seconds, we consider it a failure. The number of robust samples out of $100$ is the verification accuracy, which serves as the evaluation metric in our experiments.

\paragraph{Evaluation details of the speech recognition task.} For speech recognition, the original signal is split into several frames, followed by three preprocessing steps, including pre-emphasizing and windowing, the power spectrum of Fast Fourier transform (FFT), and the Mel-filter bank log energy. The nonlinear functions (e.g., log and square) used during preprocessing are considered in the model verification. The magnitude of signal perturbation is measured in decibels (dB), where a smaller dB value indicates a weaker perturbation.

\paragraph{Evaluation details of the sentiment analysis task.} 
We use the GloVe model to map the words into embeddings. GloVe encodes word meanings based on global co-occurrence statistics, positioning semantically similar words close together in the embedding space. Therefore, $L_\infty$-norm perturbations still make sense for the text classification task.

\section{Appendix E: Detailed Results of RQ1}
\begin{figure*}[!htbp]
    \centering
    \includegraphics[width=\textwidth]{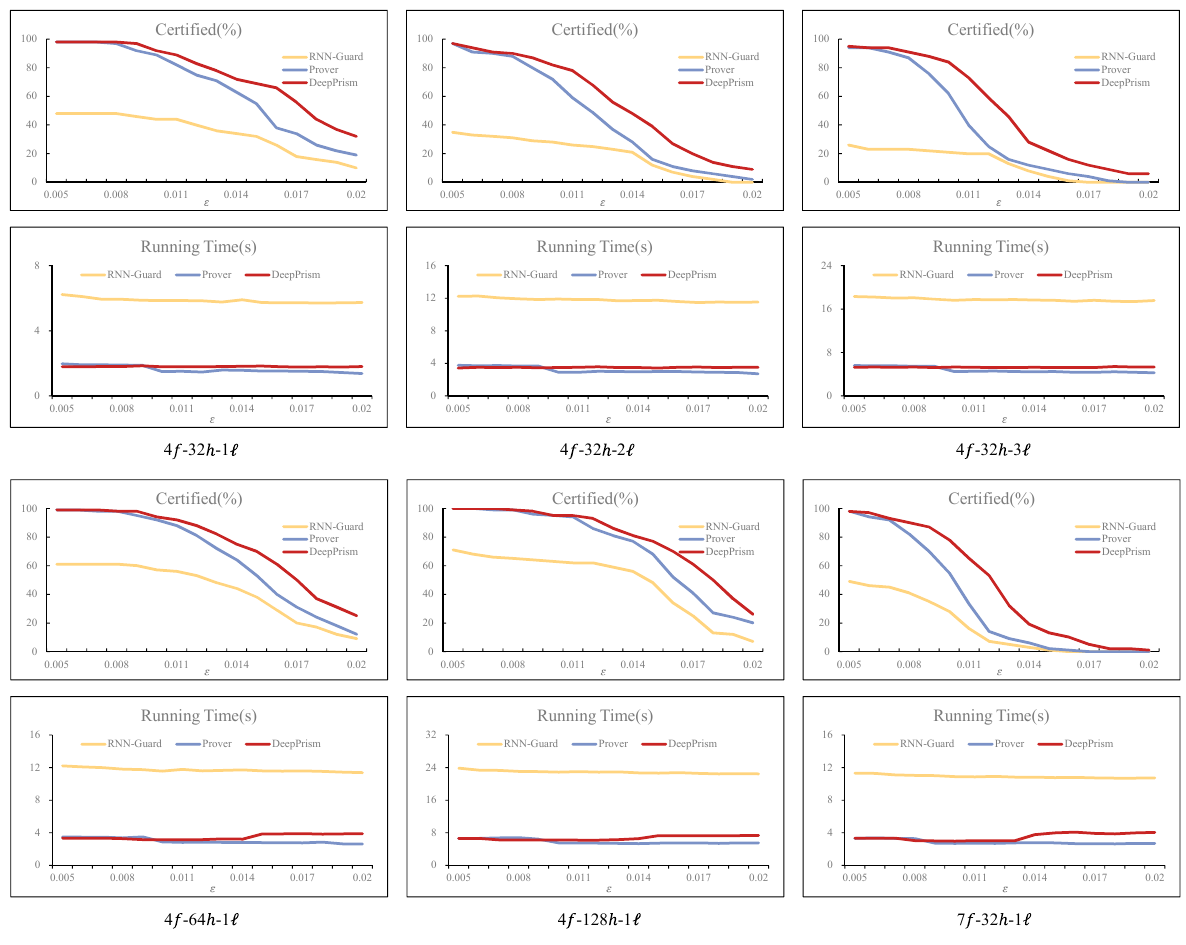}
    \caption{Results on MNIST with different perturbations and models. \emph{DeepPrism}(red solid line) the highest certified accuracy and short running time in all models.}
    \label{res-1}
\end{figure*}

The performance comparison of \emph{RNN-Guard}, \emph{Prover}, and \emph{DeepPrism} under different model parameters is shown in Figure~\ref{res-1}. Horizontally, \emph{DeepPrism} outperforms other baselines on accuracy with a slight but acceptable increase in computation time. Vertically, an increase in $f$ and $\ell$ leads to a decline in the accuracy of the verifier, while an increase in $h$ causes an improvement.

\section{Appendix F: Detailed Results of RQ2}
\begin{figure*}[!htbp]
    \centering
    \includegraphics[width=\textwidth]{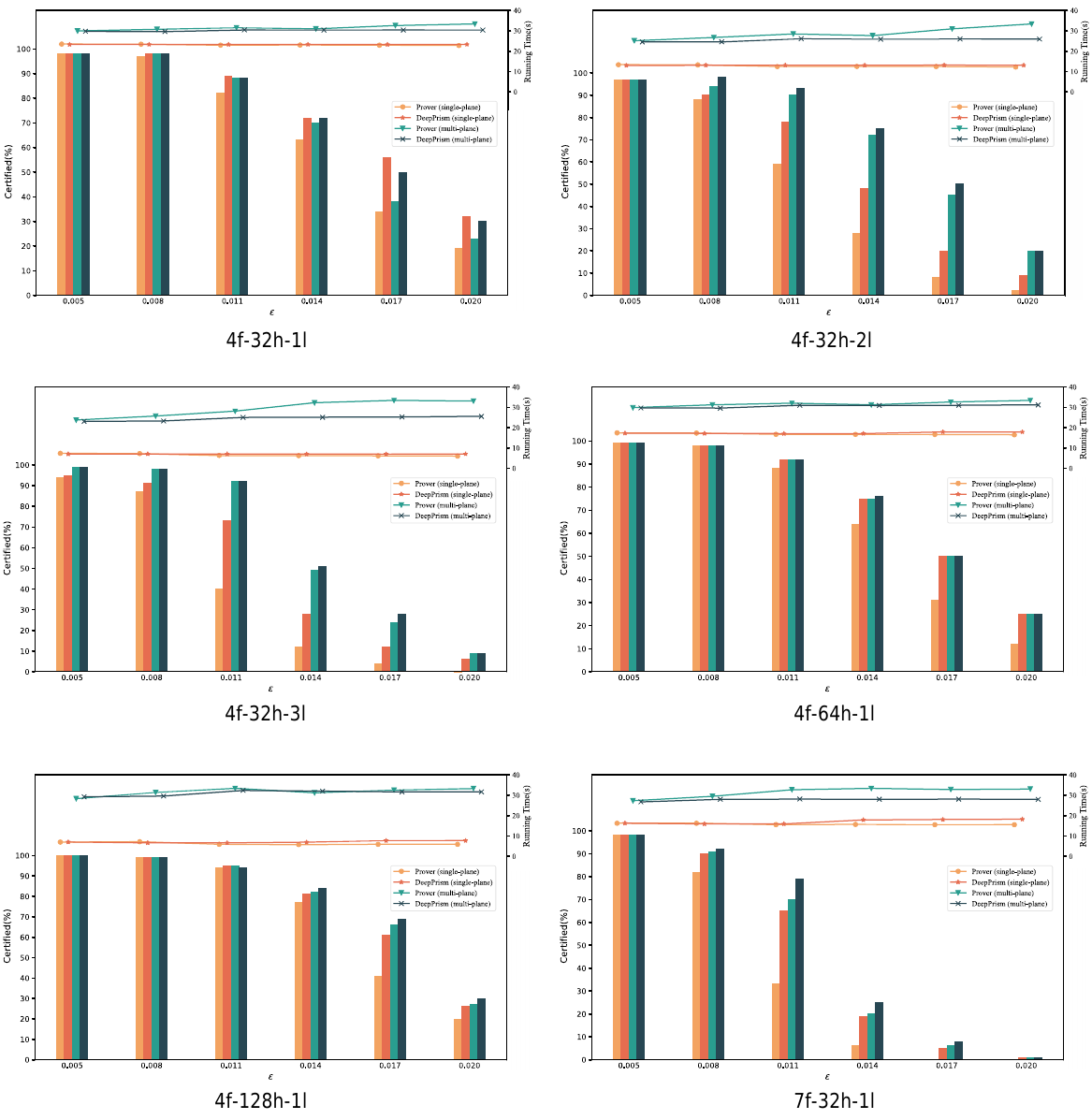}
    \caption{Comparison of two verification methods (single-plane, multi-plane) on two models (\emph{Prover} and \emph{DeepPrism}) evaluated on MNIST. Certified accuracy (bar) and running time (line) are shown in the same plot.  Experimental results show that the multi-plane method significantly outperforms the single-plane method. Under the multi-plane setting, \emph{DeepPrism} surpasses \emph{Prover}, with the performance gap widening under larger perturbations.}
    \label{res2a}
\end{figure*}

\begin{figure*}[!htbp]
    \centering
    \includegraphics[width=0.6\textwidth, height=0.3\textwidth]{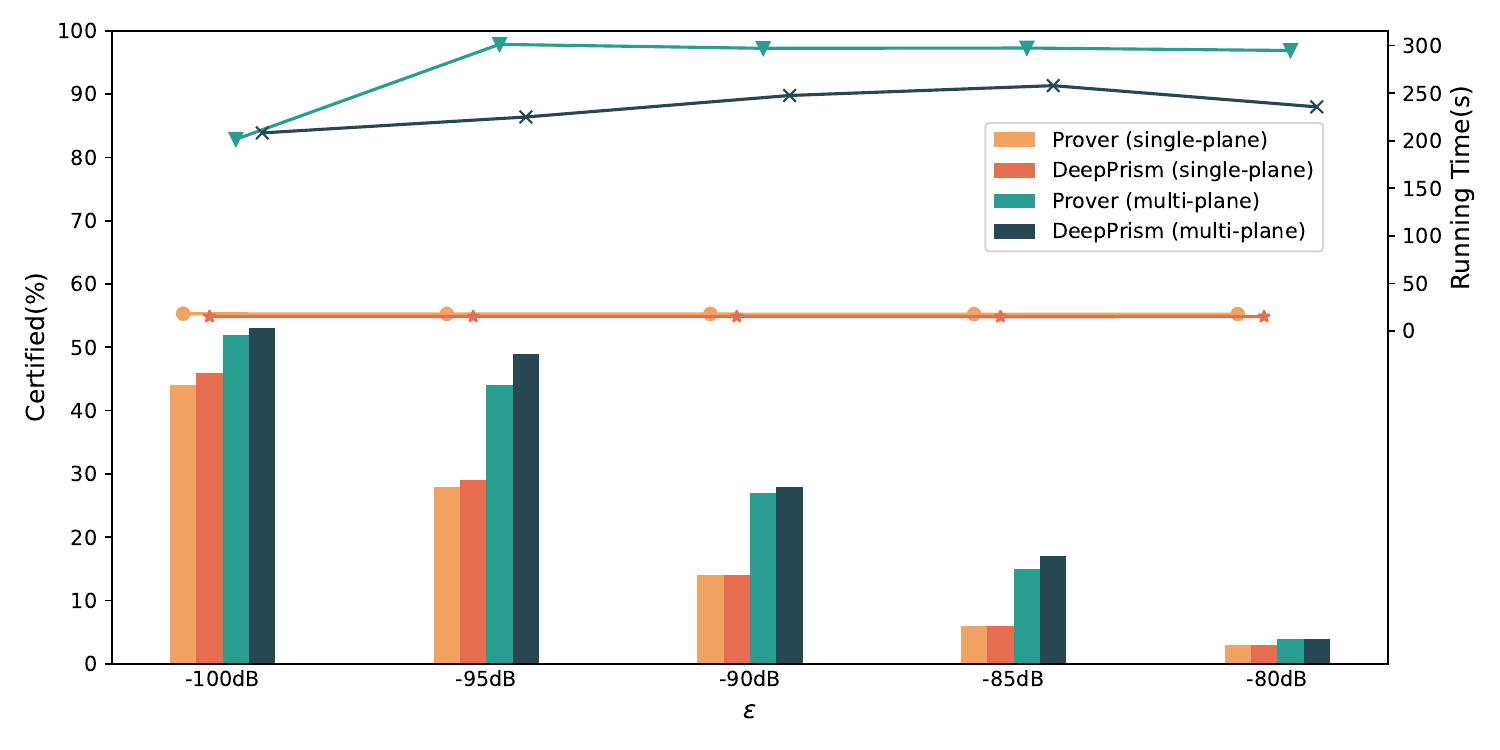}
    \caption{Comparison of two verification methods (single-plane, multi-plane) on two models (\emph{Prover} and \emph{DeepPrism}) evaluated on GSC. Certified accuracy (bar) and running time (line) are shown in the same plot.  \emph{DeepPrism} with the multi-plane method outperforms other approaches in certified accuracy.}
    \label{res2b}
\end{figure*}

\begin{figure*}[!htbp]
    \centering
    \includegraphics[width=0.6\textwidth, height=0.3\textwidth]{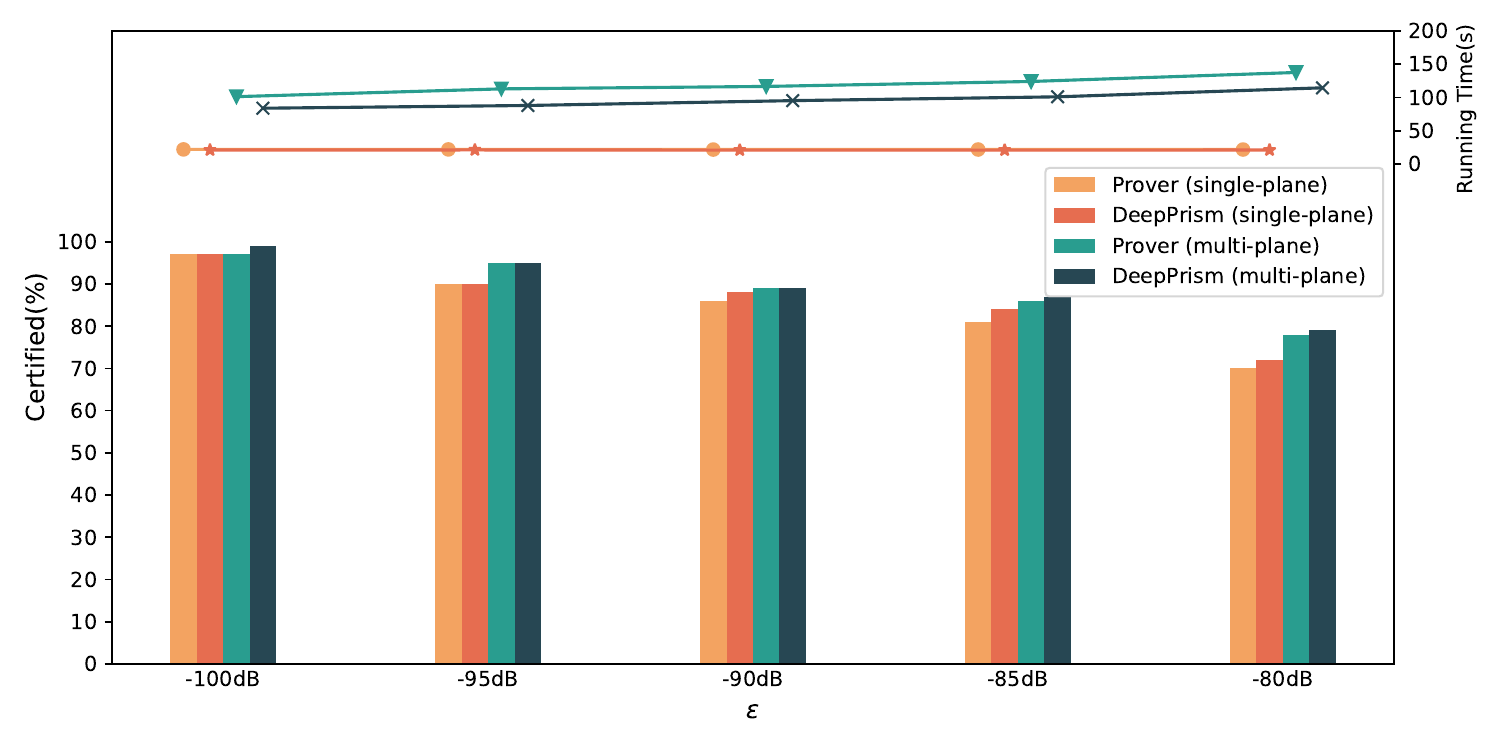}
    \caption{Comparison of two verification methods (single-plane, multi-plane) on two models (\emph{Prover} and \emph{DeepPrism}) evaluated on FDSS. Certified accuracy (bar) and running time (line) are shown in the same plot. \emph{DeepPrism} with the multi-plane method outperforms other approaches in certified accuracy.}
    \label{res2c}
\end{figure*}

\begin{figure*}[!htbp]
    \centering
    \includegraphics[width=0.6\textwidth, height=0.3\textwidth]{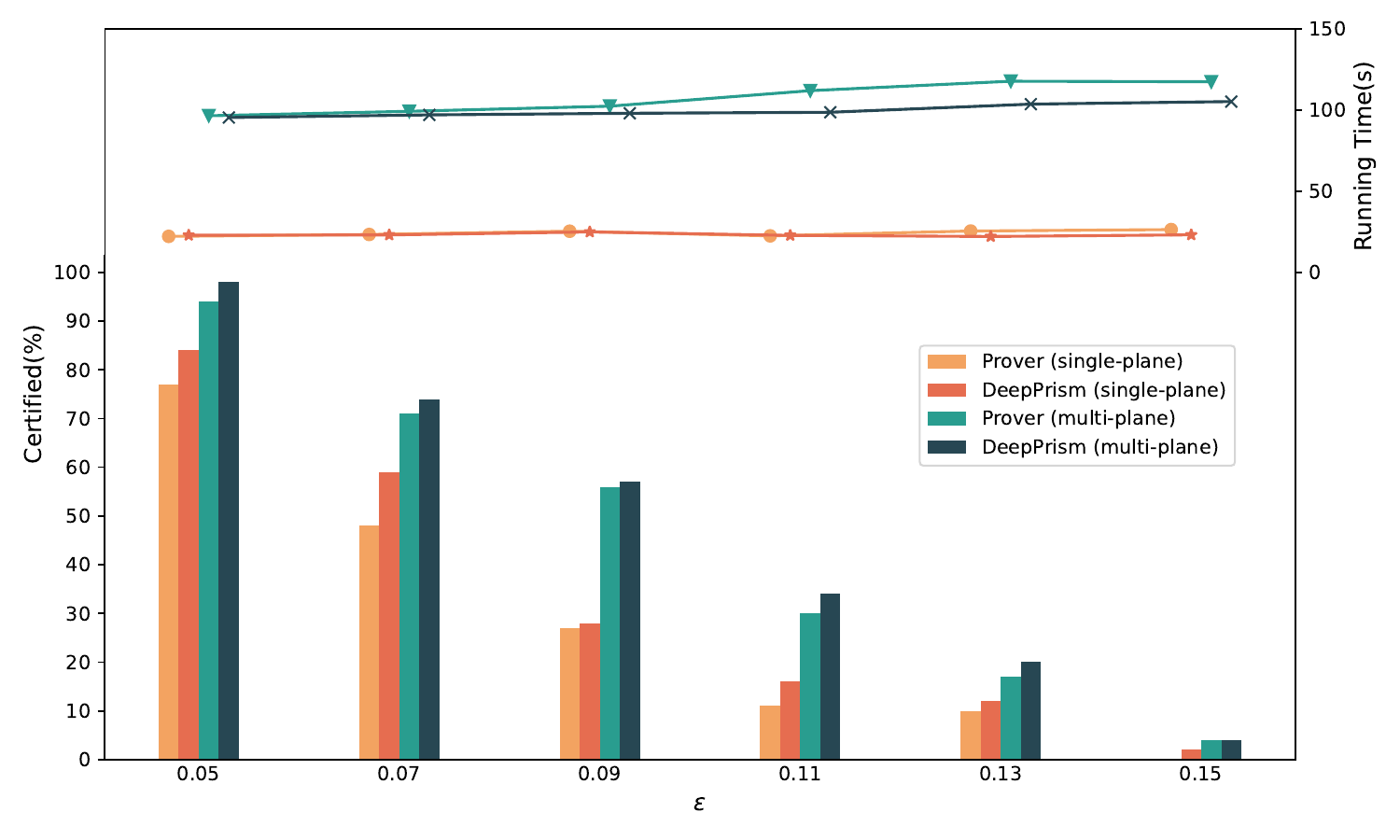}
    \caption{Comparison of two verification methods (single-plane, multi-plane) on two models (\emph{Prover} and \emph{DeepPrism}) evaluated on RT. Certified accuracy (bar) and running time (line) are shown in the same plot. \emph{DeepPrism} with the multi-plane method outperforms other approaches in certified accuracy.}
    \label{res2d}
\end{figure*}

The comparison between the single-plane and multi-plane verification methods across four datasets can be found in Figure~\ref{res2a} (MNIST, image classification), Figure~\ref{res2b} (GSC, speech recognition), Figure~\ref{res2c} (FSDD, speech recognition) and Figure~\ref{res2d} (RT, sentiment analysis), respectively. The results of multi-plane approximation are better than those of single-plane approximation in all models, and the running time increases as the accuracy decreases because verification failures require iterating all epochs.

\section{Appendix G: Detailed Results of RQ3}
The comparison between different divisons across four datasets can be found in Table~\ref{res5a}--{}\ref{res5f} (MNIST, image classification), Table~\ref{ressr1app} (GSC, speech recognition), Table~\ref{ressr2app} (FSDD, speech recognition), Table~\ref{ressaapp} (RT, sentiment analysis), respectively. Under low perturbations, all models perform similarly under all divisions, where division \textcircled{\raisebox{-0.9pt}{3}} 4-tri and \textcircled{\raisebox{-0.9pt}{6}} 4-rec slightly outperform others. Under high perturbations, division \textcircled{\raisebox{-0.9pt}{6}} 4-rec clearly outperforms others, maintaining higher accuracy. In the cases of \textcircled{\raisebox{-0.9pt}{7}} 9-rec and \textcircled{\raisebox{-0.9pt}{8}} 16-rec divisions, the verification accuracy has a significant improvement compared to that of \textcircled{\raisebox{-0.9pt}{6}} 4-rec division.

\setlength{\tabcolsep}{3pt}
\begin{table*}[hb]
\centering
\begin{tabular}{@{}c|lr|lr|lr|lr|lr|lr|lr|lr@{}}
    \toprule
     \multicolumn{1}{c|}{ } & \multicolumn{2}{c|}{\textcircled{\raisebox{-0.9pt}{1}} 2-tri-up} & \multicolumn{2}{c|}{\textcircled{\raisebox{-0.9pt}{2}} 2-tri-down} & \multicolumn{2}{c|}{\textcircled{\raisebox{-0.9pt}{3}} 4-tri} & \multicolumn{2}{c|}{\textcircled{\raisebox{-0.9pt}{4}} 2-rec-vec} & \multicolumn{2}{c|}{\textcircled{\raisebox{-0.9pt}{5}} 2-rec-hor} & \multicolumn{2}{c|}{\textcircled{\raisebox{-0.9pt}{6}} 4-rec} & \multicolumn{2}{c|}{\textcircled{\raisebox{-0.9pt}{7}} 9-rec} & \multicolumn{2}{c}{\textcircled{\raisebox{-0.9pt}{8}} 16-rec} \\
     \midrule
    $\epsilon$ & Acc. & Time & Acc. & Time & Acc. & Time & Acc. & Time & Acc. & Time & Acc. & Time & Acc. & Time  & Acc. & Time \\
    \midrule
    0.005 & \textbf{98} & 3.82  & \textbf{98} & 16.72  & \textbf{98} & 7.28  & \textbf{98} & 5.20 & \textbf{98} & 5.60  & \textbf{98} & 9.21& \textbf{98} & 17.71 & \textbf{98} & 23.94\\
    0.008 & \textbf{98} & 3.94  & \textbf{98} & 16.71  & \textbf{98} & 7.95  & \textbf{98} & 5.43  & \textbf{98} & 5.34  & \textbf{98} & 9.21 & \textbf{98} & 20.39 & \textbf{98} & 24.38\\
    0.011 & 87 & 4.83  & 88 & 19.18  & 88 & 8.59  & 85 & 6.03  & 87 & 5.84  & 88 & 9.40 & \textbf{89} & 20.90 & \textbf{89} & 24.63\\
    0.014 & 70 & 5.42  & 71 & 16.07  & 70 & 8.19 & 68 & 8.20  & 71 & 7.42  & 70 & 9.67 & \textbf{72} & 21.77 & \textbf{72} & 24.85\\
    0.017 & 37 & 6.00  & 37 & 16.10  & \textbf{38} & 9.46  & 37 & 8.85  & 37 & 8.02  & \textbf{38} & 11.62 & \textbf{38} & 23.38 & \textbf{38} & 24.98\\
    $0.020$ & 21 & 6.94  & 21 & 10.06  & \textbf{23} & 10.11  & 21 & 8.46  & 21 & 8.37 & \textbf{23} & 10.19 & \textbf{23} & 22.41 & \textbf{23} & 25.12\\
    \bottomrule
\end{tabular}
\caption{Verification accuracy of different divisons on MNIST under different perturbations where $f=4, h=32$ and $\ell=1$ .}
\label{res5a}
\end{table*}

\setlength{\tabcolsep}{3pt}
\begin{table*}[hb]
\centering
\begin{tabular}{@{}c|lr|lr|lr|lr|lr|lr|lr|lr@{}}
    \toprule
     \multicolumn{1}{c|}{ } & \multicolumn{2}{c|}{\textcircled{\raisebox{-0.9pt}{1}} 2-tri-up} & \multicolumn{2}{c|}{\textcircled{\raisebox{-0.9pt}{2}} 2-tri-down} & \multicolumn{2}{c|}{\textcircled{\raisebox{-0.9pt}{3}} 4-tri} & \multicolumn{2}{c|}{\textcircled{\raisebox{-0.9pt}{4}} 2-rec-vec} & \multicolumn{2}{c|}{\textcircled{\raisebox{-0.9pt}{5}} 2-rec-hor} & \multicolumn{2}{c|}{\textcircled{\raisebox{-0.9pt}{6}} 4-rec} & \multicolumn{2}{c|}{\textcircled{\raisebox{-0.9pt}{7}} 9-rec} & \multicolumn{2}{c}{\textcircled{\raisebox{-0.9pt}{8}} 16-rec} \\
     \midrule
    $\epsilon$ & Acc. & Time & Acc. & Time & Acc. & Time & Acc. & Time & Acc. & Time & Acc. & Time & Acc. & Time  & Acc. & Time \\
    \midrule
    0.005 & \textbf{97} & 7.79  & \textbf{97} & 7.80  & \textbf{97} & 14.58  & \textbf{97} & 10.81  & \textbf{97} & 8.42  & \textbf{97} & 18.69  & \textbf{97} & 28.18 & \textbf{97} & 38.02\\
    0.008 & 94 & 8.82  & 94 & 9.14  & 94 & 15.97  & 93 & 11.17  & 94 & 10.93  & 95 & 18.98  & \textbf{97} & 28.40  & \textbf{97} & 37.85\\
    0.011 & 88 & 9.84  & 87 & 9.55  & 90 & 17.56  & 81 & 12.42  & 95 & 11.55  & \textbf{97} & 19.00  & \textbf{97} & 29.89  & \textbf{97} & 37.85\\
    0.014 & 68 & 12.13  & 72 & 13.68  & 72 & 16.87  & 54 & 19.62  & 86 & 17.26  & 93 & 19.74  & 95 & 33.09  & \textbf{96} & 38.69\\
    0.017 & 44 & 13.08  & 39 & 16.88  & 45 & 19.86  & 23 & 19.17  & 64 & 18.98  & 65 & 21.42  & 71 & 33.15  & \textbf{76} & 37.64\\
    $0.020$ & 18 & 14.47  & 18 & 21.91  & 20 & 22.07  & 7 & 17.06  & 28 & 19.53  & 54 & 21.72  & 57 & 31.68  & \textbf{64} & 38.39\\
    \bottomrule
\end{tabular}
\caption{Verification accuracy of different divisons on MNIST under different perturbations where $f=4, h=32$ and $\ell=2$ .}
\label{res5b}
\end{table*}

\setlength{\tabcolsep}{3pt}
\begin{table*}[ht]
\centering
\begin{tabular}{@{}c|lr|lr|lr|lr|lr|lr|lr|lr@{}}
    \toprule
     \multicolumn{1}{c|}{ } & \multicolumn{2}{c|}{\textcircled{\raisebox{-0.9pt}{1}} 2-tri-up} & \multicolumn{2}{c|}{\textcircled{\raisebox{-0.9pt}{2}} 2-tri-down} & \multicolumn{2}{c|}{\textcircled{\raisebox{-0.9pt}{3}} 4-tri} & \multicolumn{2}{c|}{\textcircled{\raisebox{-0.9pt}{4}} 2-rec-vec} & \multicolumn{2}{c|}{\textcircled{\raisebox{-0.9pt}{5}} 2-rec-hor} & \multicolumn{2}{c|}{\textcircled{\raisebox{-0.9pt}{6}} 4-rec} & \multicolumn{2}{c|}{\textcircled{\raisebox{-0.9pt}{7}} 9-rec} & \multicolumn{2}{c}{\textcircled{\raisebox{-0.9pt}{8}} 16-rec} \\
     \midrule
    $\epsilon$ & Acc. & Time & Acc. & Time & Acc. & Time & Acc. & Time & Acc. & Time & Acc. & Time & Acc. & Time  & Acc. & Time \\
    \midrule
    0.005 & \textbf{99} & 11.70  & \textbf{99} & 50.08  & \textbf{99} & 21.70  & \textbf{99} & 15.24  & \textbf{99} & 14.76  & \textbf{99} & 28.18 & \textbf{99} & 60.13 & \textbf{99} & 71.71\\
    0.008 & 97 & 13.55  & 98 & 51.99  & 98 & 23.65  & 97 & 16.34  & \textbf{99} & 15.71  & \textbf{99} & 28.40 & \textbf{99} & 56.24 & \textbf{99} & 72.26\\
    0.011 & 83 & 16.37  & 82 & 68.98  & 92 & 25.95  & 63 & 21.59  & 89 & 19.27  & \textbf{95} & 29.89 & \textbf{95} & 64.10 & \textbf{95} & 71.91\\
    0.014 & 34 & 19.82  & 33 & 23.66  & 49 & 30.10  & 25 & 27.78  & 46 & 31.83  & 71 & 33.09 & 80 & 72.79 & \textbf{83} & 72.46\\
    0.017 & 16 & 19.26  & 14 & 24.22  & 24 & 31.15  & 7 & 26.63  & 21 & 28.10  & 33 & 33.15 & 35 & 65.34 & \textbf{46} & 71.83\\
    $0.020$ & 1 & 21.44  & 1 & 31.40  & 9 & 30.90  & 0 & 24.06  & 5 & 26.09  & \textbf{15} & 31.68 & \textbf{15} & 73.36 & \textbf{15} & 72.13\\
    \bottomrule
\end{tabular}
\caption{Verification accuracy of different divisons on MNIST under different perturbations where $f=4, h=32$ and $\ell=3$ .}
\label{res5c}
\end{table*}

\setlength{\tabcolsep}{3pt}
\begin{table*}[ht]
\centering
\begin{tabular}{@{}c|lr|lr|lr|lr|lr|lr|lr|lr@{}}
    \toprule
     \multicolumn{1}{c|}{ } & \multicolumn{2}{c|}{\textcircled{\raisebox{-0.9pt}{1}} 2-tri-up} & \multicolumn{2}{c|}{\textcircled{\raisebox{-0.9pt}{2}} 2-tri-down} & \multicolumn{2}{c|}{\textcircled{\raisebox{-0.9pt}{3}} 4-tri} & \multicolumn{2}{c|}{\textcircled{\raisebox{-0.9pt}{4}} 2-rec-vec} & \multicolumn{2}{c|}{\textcircled{\raisebox{-0.9pt}{5}} 2-rec-hor} & \multicolumn{2}{c|}{\textcircled{\raisebox{-0.9pt}{6}} 4-rec} & \multicolumn{2}{c|}{\textcircled{\raisebox{-0.9pt}{7}} 9-rec} & \multicolumn{2}{c}{\textcircled{\raisebox{-0.9pt}{8}} 16-rec} \\
     \midrule
    $\epsilon$ & Acc. & Time & Acc. & Time & Acc. & Time & Acc. & Time & Acc. & Time & Acc. & Time & Acc. & Time  & Acc. & Time \\
    \midrule
    0.005 & \textbf{99} & 7.52  & \textbf{99} & 10.87  & \textbf{99} & 14.40  & \textbf{99} & 10.38  & \textbf{99} & 9.88  & \textbf{99} & 19.09 & \textbf{99} & 40.04 & \textbf{99} & 48.26\\
    0.008 & \textbf{98} & 8.50  & \textbf{98} & 10.62  & \textbf{98} & 15.63  & \textbf{98} & 10.15  & \textbf{98} & 9.92  & \textbf{98} & 19.01 & \textbf{98} & 40.30 & \textbf{98} & 48.56\\
    0.011 & 91 & 8.87  & 91 & 11.28  & \textbf{92} & 16.28  & 91 & 10.91  & 91 & 10.73  & \textbf{92} & 19.19 & \textbf{92} & 37.60 & \textbf{92} & 48.37\\
    0.014 & 71 & 9.56  & 71 & 10.86  & 75 & 15.61  & 68 & 12.87  & 71 & 13.07  & 78 & 20.90 & 80 & 42.55 & \textbf{81} & 48.94\\
    0.017 & 33 & 10.15  & 33 & 12.97  & 50 & 16.78  & 33 & 13.73  & 33 & 14.01  & 58 & 19.39 & 59 & 39.81 & \textbf{62} & 48.89\\
    $0.020$ & 18 & 11.25  & 18 & 15.58  & 25 & 17.49  & 16 & 13.22  & 18 & 12.95  & 28 & 19.72 & 34 & 44.14 & \textbf{38} & 50.42\\
    \bottomrule
\end{tabular}
\caption{Verification accuracy of different divisons on MNIST under different perturbations where $f=4, h=64$ and $\ell=1$ .}
\label{res5d}
\end{table*}

\setlength{\tabcolsep}{3pt}
\begin{table*}[ht]
\centering
\begin{tabular}{@{}c|lr|lr|lr|lr|lr|lr|lr|lr@{}}
    \toprule
     \multicolumn{1}{c|}{ } & \multicolumn{2}{c|}{\textcircled{\raisebox{-0.9pt}{1}} 2-tri-up} & \multicolumn{2}{c|}{\textcircled{\raisebox{-0.9pt}{2}} 2-tri-down} & \multicolumn{2}{c|}{\textcircled{\raisebox{-0.9pt}{3}} 4-tri} & \multicolumn{2}{c|}{\textcircled{\raisebox{-0.9pt}{4}} 2-rec-vec} & \multicolumn{2}{c|}{\textcircled{\raisebox{-0.9pt}{5}} 2-rec-hor} & \multicolumn{2}{c|}{\textcircled{\raisebox{-0.9pt}{6}} 4-rec} & \multicolumn{2}{c|}{\textcircled{\raisebox{-0.9pt}{7}} 9-rec} & \multicolumn{2}{c}{\textcircled{\raisebox{-0.9pt}{8}} 16-rec} \\
     \midrule
    $\epsilon$ & Acc. & Time & Acc. & Time & Acc. & Time & Acc. & Time & Acc. & Time & Acc. & Time & Acc. & Time  & Acc. & Time \\
    \midrule
    0.005 & \textbf{100} & 15.01  & \textbf{100} & 18.81 & \textbf{100} & 27.90  & \textbf{100} & 21.51 & \textbf{100} & 17.52 & \textbf{100} & 38.02 & \textbf{100} & 80.22 & \textbf{100} & 96.43\\
    0.008 & \textbf{99} & 16.43 & \textbf{99} & 18.84 & \textbf{99} & 30.89 & \textbf{99} & 19.70 & \textbf{99} & 18.99 & \textbf{99} & 37.85 & \textbf{99} & 77.75 & \textbf{99} & 97.53\\
    0.011 & 94 & 16.94  & 94 & 17.74  & \textbf{95} & 32.86  & 94 & 21.38  & 94 & 20.26  & \textbf{95} & 37.85 & \textbf{95} & 71.62 & \textbf{95} & 96.38\\
    0.014 & 78 & 17.76  & 78 & 20.15  & 82 & 30.74  & 77 & 24.43  & 78 & 23.04  & 83 & 38.69 & 87 & 83.78 & \textbf{88} & 95.85\\
    0.017 & 50 & 18.27  & 50 & 23.14  & 66 & 32.05  & 47 & 25.23  & 50 & 24.01  & 66 & 37.64 & \textbf{69} & 75.57 & \textbf{69} & 97.12\\
    $0.020$ & 22 & 20.63  & 22 & 26.45  & 27 & 32.83  & 21 & 22.32  & 22 & 23.37  & 27 & 38.39 & 34 & 80.95 & \textbf{38} & 96.56\\
    \bottomrule
\end{tabular}
\caption{Verification accuracy of different divisons on MNIST under different perturbations where $f=4, h=128$ and $\ell=1$ .}
\label{res5e}
\end{table*}

\setlength{\tabcolsep}{3pt}
\begin{table*}[ht]
\centering
\begin{tabular}{@{}c|lr|lr|lr|lr|lr|lr|lr|lr@{}}
    \toprule
     \multicolumn{1}{c|}{ } & \multicolumn{2}{c|}{\textcircled{\raisebox{-0.9pt}{1}} 2-tri-up} & \multicolumn{2}{c|}{\textcircled{\raisebox{-0.9pt}{2}} 2-tri-down} & \multicolumn{2}{c|}{\textcircled{\raisebox{-0.9pt}{3}} 4-tri} & \multicolumn{2}{c|}{\textcircled{\raisebox{-0.9pt}{4}} 2-rec-vec} & \multicolumn{2}{c|}{\textcircled{\raisebox{-0.9pt}{5}} 2-rec-hor} & \multicolumn{2}{c|}{\textcircled{\raisebox{-0.9pt}{6}} 4-rec} & \multicolumn{2}{c|}{\textcircled{\raisebox{-0.9pt}{7}} 9-rec} & \multicolumn{2}{c}{\textcircled{\raisebox{-0.9pt}{8}} 16-rec} \\
     \midrule
    $\epsilon$ & Acc. & Time & Acc. & Time & Acc. & Time & Acc. & Time & Acc. & Time & Acc. & Time & Acc. & Time  & Acc. & Time \\
    \midrule
    0.005 & \textbf{98} & 7.29 & \textbf{98} & 11.08 & \textbf{98} & 13.16 & \textbf{98} & 9.63 & \textbf{98} & 9.02 & \textbf{98} & 16.69 & \textbf{98} & 36.70 & \textbf{98} & 43.66\\
    0.008 & 88 & 8.36  & 88 & 13.08  & 91 & 15.06  & 88 & 10.69  & 88 & 10.38  & \textbf{92} & 16.96 & \textbf{92} & 37.76 & \textbf{92} & 44.92\\
    0.011 & 37 & 11.09  & 38 & 13.97  & 70 & 17.96  & 37 & 15.17  & 37 & 14.25  & \textbf{71} & 18.33 & \textbf{71} & 41.17 & \textbf{71} & 44.51\\
    0.014 & 7 & 12.78  & 7 & 14.45 & 20 & 18.50  & 7 & 18.25  & 7 & 18.24  & 27 & 19.24 & \textbf{32} & 44.45 & \textbf{32} & 45.21\\
    0.017 & 0 & 11.93  & 0 & 16.41  & 6 & 18.06  & 0 & 19.17  & 0 & 17.04  & 8 & 18.77 & \textbf{10} & 40.60 & \textbf{10} & 44.77\\
    $0.020$ & 0 & 13.10  & 0 & 19.40  & 1 & 18.26  & 1 & 13.12  & 1 & 15.31  & \textbf{3} & 18.92 & \textbf{3} & 40.76 & \textbf{3} & 45.31\\
    \bottomrule
\end{tabular}
\caption{Verification accuracy of different divisons on MNIST under different perturbations where $f=7$, $h=32$ and $\ell=1$ .}
\label{res5f}
\end{table*}

\setlength{\tabcolsep}{3pt}
\begin{table*}[ht]
\centering
\begin{tabular}{@{}c|lr|lr|lr|lr|lr|lr|lr|lr@{}}
    \toprule
     \multicolumn{1}{c|}{ } & \multicolumn{2}{c|}{\textcircled{\raisebox{-0.9pt}{1}} 2-tri-up} & \multicolumn{2}{c|}{\textcircled{\raisebox{-0.9pt}{2}} 2-tri-down} & \multicolumn{2}{c|}{\textcircled{\raisebox{-0.9pt}{3}} 4-tri} & \multicolumn{2}{c|}{\textcircled{\raisebox{-0.9pt}{4}} 2-rec-vec} & \multicolumn{2}{c|}{\textcircled{\raisebox{-0.9pt}{5}} 2-rec-hor} & \multicolumn{2}{c|}{\textcircled{\raisebox{-0.9pt}{6}} 4-rec} & \multicolumn{2}{c|}{\textcircled{\raisebox{-0.9pt}{7}} 9-rec} & \multicolumn{2}{c}{\textcircled{\raisebox{-0.9pt}{8}} 16-rec} \\
     \midrule
    $\epsilon$ & Acc. & Time & Acc. & Time & Acc. & Time & Acc. & Time & Acc. & Time & Acc. & Time & Acc. & Time  & Acc. & Time \\
    \midrule
    -100 & 31 & 213.78  & 31 & 218.91 & 52 & 201.46 & 50 & 206.33 & 51 & 210.88 & 55 & 205.84 & 64 & 435.15  & \textbf{67} & 457.99 \\
    -95 & 18 & 219.62  & 18 & 263.13  & 44 & 301.25  & 41 & 224.08  & 41 & 219.81  & 48 & 299.48 & 50 & 486.18  & \textbf{52} & 517.85 \\
    -90 & 14 & 254.91  & 13 & 216.15  & 27 & 297.09  & 35 & 232.41  & 35 & 241.06  & 29 & 303.30  & \textbf{30} & 561.35  & \textbf{30} & 579.64 \\
    -85 & 2 & 233.72  & 3 & 227.14  & 15 & 297.39  & 10 & 219.36  & 10 & 223.75  & 15 & 295.24  & 18 & 611.89  & \textbf{23} & 624.35 \\
    -80 & 0 & 286.03  & 0 & 286.77  & 4 & 294.80  & 3 & 272.17  & 3 & 298.10  & 4 & 303.92  & 4 & 644.22  & \textbf{6} & 681.95 \\
    \bottomrule
\end{tabular}
\caption{Verification accuracy of different divisons on GSC dataset under different perturbations.}
\label{ressr1app}
\end{table*}

\setlength{\tabcolsep}{3pt}
\begin{table*}[ht]
\centering
\begin{tabular}{@{}c|lr|lr|lr|lr|lr|lr|lr|lr@{}}
    \toprule
     \multicolumn{1}{c|}{ } & \multicolumn{2}{c|}{\textcircled{\raisebox{-0.9pt}{1}} 2-tri-up} & \multicolumn{2}{c|}{\textcircled{\raisebox{-0.9pt}{2}} 2-tri-down} & \multicolumn{2}{c|}{\textcircled{\raisebox{-0.9pt}{3}} 4-tri} & \multicolumn{2}{c|}{\textcircled{\raisebox{-0.9pt}{4}} 2-rec-vec} & \multicolumn{2}{c|}{\textcircled{\raisebox{-0.9pt}{5}} 2-rec-hor} & \multicolumn{2}{c|}{\textcircled{\raisebox{-0.9pt}{6}} 4-rec} & \multicolumn{2}{c|}{\textcircled{\raisebox{-0.9pt}{7}} 9-rec} & \multicolumn{2}{c}{\textcircled{\raisebox{-0.9pt}{8}} 16-rec} \\
     \midrule
    $\epsilon$ & Acc. & Time & Acc. & Time & Acc. & Time & Acc. & Time & Acc. & Time & Acc. & Time & Acc. & Time  & Acc. & Time \\
    \midrule
    -100 & \textbf{97} & 89.48 & \textbf{97} & 91.40 & \textbf{97} & 101.11 & \textbf{97} & 87.99 & \textbf{97} & 84.54 & \textbf{97} & 96.43 & \textbf{97} & 203.35 & \textbf{97} & 235.52 \\
    -95 & \textbf{95} & 94.57 & \textbf{95} & 91.81  & \textbf{95} & 112.79  & \textbf{95} & 96.64 & \textbf{95} & 94.86 & \textbf{95} & 95.66 & \textbf{95} & 207.19 & \textbf{95} & 253.08 \\
    -90 & 87 & 105.98  & 87 & 110.32  & 89 & 116.45  & 87 & 108.57 & 87 & 104.45  & 89 & 112.23  & \textbf{91} & 223.49 & \textbf{91} & 251.12 \\
    -85 & 84 & 109.01 & 85 & 108.14 & 86 & 123.89 & 85 & 110.83  & 84 & 114.50 & 86 & 118.59 & \textbf{89} & 241.20 & \textbf{89} & 289.95 \\
    -80 & 70 & 118.04  & 68 & 120.22  & 78 & 137.39  & 71 & 115.96  & 61 & 117.85  & 81 & 133.43  & 82 & 258.86 & \textbf{85} & 291.99 \\
    \bottomrule
\end{tabular}
\caption{Verification accuracy of different divisons on FSDD dataset under different perturbations.}
\label{ressr2app}
\end{table*}

\setlength{\tabcolsep}{3pt}
\begin{table*}[ht]
\centering
\begin{tabular}{@{}c|lr|lr|lr|lr|lr|lr|lr|lr@{}}
    \toprule
     \multicolumn{1}{c|}{ } & \multicolumn{2}{c|}{\textcircled{\raisebox{-0.9pt}{1}} 2-tri-up} & \multicolumn{2}{c|}{\textcircled{\raisebox{-0.9pt}{2}} 2-tri-down} & \multicolumn{2}{c|}{\textcircled{\raisebox{-0.9pt}{3}} 4-tri} & \multicolumn{2}{c|}{\textcircled{\raisebox{-0.9pt}{4}} 2-rec-vec} & \multicolumn{2}{c|}{\textcircled{\raisebox{-0.9pt}{5}} 2-rec-hor} & \multicolumn{2}{c|}{\textcircled{\raisebox{-0.9pt}{6}} 4-rec} & \multicolumn{2}{c|}{\textcircled{\raisebox{-0.9pt}{7}} 9-rec} & \multicolumn{2}{c}{\textcircled{\raisebox{-0.9pt}{8}} 16-rec} \\
     \midrule
    $\epsilon$ & Acc. & Time & Acc. & Time & Acc. & Time & Acc. & Time & Acc. & Time & Acc. & Time & Acc. & Time  & Acc. & Time \\
    \midrule
    0.05 & 94 & 81.73 & 94 & 74.41 & 94 & 96.40 & 94 & 73.23  & 94 & 71.89  & \textbf{95} & 93.19  & \textbf{95} & 181.42  & \textbf{95} & 222.67 \\
    0.07 & 68 & 78.59  & 64 & 74.50  & 71 & 99.14  & 70 & 77.38  & 70 & 77.95  & 78 & 94.47  & 82 & 189.20  & \textbf{88} & 233.38 \\
    0.09 & 39 & 72.92  & 42 & 83.98  & 56 & 102.29  & 48 & 81.56  & 49 & 80.12  & 61 & 104.13  & 66 & 205.32  & \textbf{69} & 253.22 \\
    0.11 & 21 & 93.63  & 18 & 86.26  & 30 & 111.85  & 21 & 86.12  & 21 & 87.90  & 39 & 113.60  & 46 & 222.91  & \textbf{54} & 273.01 \\
    0.13 & 10 & 89.78  & 10 & 90.65  & 17 & 117.66  & 10 & 88.75  & 9 & 86.03  & 19 & 114.73  & 22 & 227.63  & \textbf{28} & 281.49 \\
    0.15 & 0 & 96.15  & 0 & 92.01  & \textbf{4} & 117.34  & 1 & 92.34  & 0 & 94.76  & \textbf{4} & 116.32  & \textbf{4} & 236.38  & \textbf{4} & 293.80 \\

    \bottomrule
\end{tabular}
\caption{Verification accuracy of different divisons on RT dataset under different perturbations.}
\label{ressaapp}
\end{table*}

\end{document}